\documentclass[10pt,twocolumn,letterpaper]{article}

\usepackage[pagenumbers]{cvpr} % To force page numbers, e.g. for an arXiv version

\usepackage{graphicx}
\usepackage{times}
\usepackage{helvet}
\usepackage{courier}
\usepackage{amsmath}
\usepackage{algorithm}
\usepackage{algorithmic}
\usepackage{csquotes} 
\usepackage{color}
\usepackage{paralist}
\usepackage{amssymb}
\usepackage{indentfirst}
\usepackage{pifont}
\usepackage{float}
\usepackage{multirow}
\usepackage{mathrsfs}
\usepackage{makecell}

\usepackage{mathrsfs}
\usepackage{textcomp,booktabs}
\usepackage{amssymb}% http://ctan.org/pkg/amssymb
\usepackage{pifont}% http://ctan.org/pkg/pifont
\usepackage[misc]{ifsym}
\newcommand{\cmark}{\ding{51}}%
\newcommand{\xmark}{\ding{55}}%

\usepackage{CJKutf8}

\usepackage{soul}
\usepackage{colortbl}
\usepackage[dvipsnames]{xcolor}
\usepackage{mathrsfs}
\usepackage{color}
\usepackage{xcolor}
\definecolor{citecolor}{HTML}{0071bc} 
\definecolor{SeaGreen4}{RGB}{0,205,102} 
\definecolor{SlateBlue}{RGB}{106,90,205} 
\definecolor{DarkRed}{RGB}{178,34,34} 
\usepackage[colorlinks, linkcolor=red,  anchorcolor=blue, citecolor=citecolor]{hyperref}

\usepackage{colortbl}
\definecolor{mygray}{gray}{.9}
\definecolor{mypink}{rgb}{.99,.91,.95}
\definecolor{mycyan}{cmyk}{.3,0,0,0}

\usepackage{color}
\usepackage{xcolor}
\definecolor{citecolor}{HTML}{0071bc} 
\definecolor{SeaGreen4}{RGB}{0,205,102} 
\definecolor{SlateBlue}{RGB}{106,90,205} 
\definecolor{DarkRed}{RGB}{178,34,34}

\usepackage[capitalize]{cleveref}
\crefname{section}{Sec.}{Secs.}
\Crefname{section}{Section}{Sections}
\Crefname{table}{Table}{Tables}
\crefname{table}{Tab.}{Tabs.}

\def\paperID{*****} % *** Enter the Paper ID here
\def\confName{CVPR}
\def\confYear{2026}

\title{ Event Stream-based Sign Language Translation: A High-Definition Benchmark Dataset and A Novel Baseline }  

\author{Shiao Wang$^{1}$, Xiao Wang$^{1}$\thanks{\Letter~Corresponding Author: Xiao Wang}, Duoqing Yang$^{1}$, Yao Rong$^{1}$, 
        Fuling Wang$^{1}$, Jianing Li$^{2}$, Lin Zhu$^{3}$, Bo Jiang$^{1}$ \\ 
${^1}${School of Computer Science and Technology, Anhui University, Hefei, China} \\
${^2}${Qiyuan Laboratory, Beijing, China} \\
${^3}${Beijing Institute of Technology, Beijing, China} \\
\textit{e24101001@stu.ahu.edu.cn}, \textit{xiaowang@ahu.edu.cn},  \textit{e125221163@stu.ahu.edu.cn}, \\ 
\textit{rytaptap@163.com}, \textit{e23201049@stu.ahu.edu.cn}, \textit{}, \\
\textit{linzhu@pku.edu.cn},  \textit{jiangbo@ahu.edu.cn} 
}
\begin{document}
\maketitle

%%%%%%%%% ABSTRACT
% \begin{abstract}
% Sign Language Translation (SLT) is a core task in the field of AI-assisted disability. Unlike traditional SLT based on visible light videos, which is easily affected by factors such as lighting, rapid hand movements, and privacy breaches, this paper proposes the use of high-definition Event streams for SLT, effectively mitigating the aforementioned issues. This is primarily because Event streams have a high dynamic range and dense temporal signals, which can withstand low illumination and motion blur well. Additionally, due to their sparsity in space, they effectively protect the privacy of the target person. More specifically, we propose a new high-definition Event stream sign language dataset, termed Event-CSL, which effectively fills the data gap in this area of research. It contains 14,827 videos, 14,821 glosses, and 2,544 Chinese words in the text vocabulary. These samples are collected in a variety of indoor and outdoor scenes, encompassing multiple angles, light intensities, and camera movements. We have benchmarked existing mainstream SLT works to enable fair comparison for future efforts. Based on this dataset and several other large-scale datasets, we propose a novel baseline method that fully leverages the Mamba model's ability to aggregate temporal information of CNN features, resulting in improved sign language translation outcomes. Both the benchmark dataset and source code will be released. 
% \end{abstract} 

\begin{abstract}
Sign Language Translation (SLT) is a core task in the field of AI-assisted disability. Traditional SLT methods are typically based on visible light videos, which are easily affected by factors such as lighting variations, rapid hand movements, and privacy concerns. This paper proposes the use of bio-inspired event cameras to alleviate the aforementioned issues. Specifically, we introduce a new high-definition event-based sign language dataset, termed Event-CSL, which effectively addresses the data scarcity in this research area. The dataset comprises 14,827 videos, 14,821 glosses, and 2,544 Chinese words in the text vocabulary. These samples are collected across diverse indoor and outdoor scenes, covering multiple viewpoints, lighting conditions, and camera motions. We have also benchmarked existing mainstream SLT methods on this dataset to facilitate fair comparisons in future research.
Furthermore, we propose a novel event-based sign language translation framework, termed EvSLT. The framework first segments continuous video features into clips and employs a Mamba-based memory aggregation module to compress and aggregate spatial detail features at the clip level. Subsequently, these spatial features, along with temporal representations obtained from temporal convolution, are then fused by a graph-guided spatiotemporal fusion module. Extensive experiments on Event-CSL, as well as other publicly available datasets, demonstrate the superior performance of our method. The dataset and source code will be released on \url{https://github.com/Event-AHU/OpenESL} 
\end{abstract}

\section{Introduction} 

With the rapid development of deep learning, \textit{AI (Artificial Intelligence) for good} has received widespread attention, among which, Sign Language Translation (SLT)~\cite{ye2023cross, yin2023gloss, Zhou_2023_ICCV} is increasingly being emphasized. It facilitates communication between deaf individuals and non-signers by translating sign language videos into natural language text, thereby significantly enhancing social participation and quality of life for the deaf community. However, the performance of sign language translation is still limited due to the usage of traditional RGB cameras, as it is easily influenced by limited frame rate, illumination, motion blur, etc. In addition, the deployment of RGB-based sign language translation models raises potential ethical concerns, as it inherently involves issues related to human privacy.

Recently, biologically inspired event cameras (e.g., DVS346, Prophesee, CeleX) have drawn more and more attention due to their unique advantages in \textit{high dynamic range}, \textit{low energy cost}, \textit{sparse spatial but dense temporal resolution}, and \textit{low latency}, etc.~\cite {gallego2020eventSurvey}. In terms of the imaging mechanism, each pixel in an event camera operates independently and asynchronously, which fundamentally differs from RGB cameras that capture synchronized frame-based outputs. A binary pulse signal is recorded only when the change in brightness for each pixel exceeds a specific threshold. Some computer vision tasks have involved the event camera for event-based or event-enhanced learning, such as object detection~\cite{gehrig2024lowlatencyDetEvent}, tracking~\cite{wang2024eventvot, wang2024visevent}, action recognition~\cite{wang2024hardvs, wang2021eventgaits}, and also the sign language translation~\cite{zhang2024evsign} discussed in this paper.

\begin{figure}
\centering
\includegraphics[width=1\linewidth]{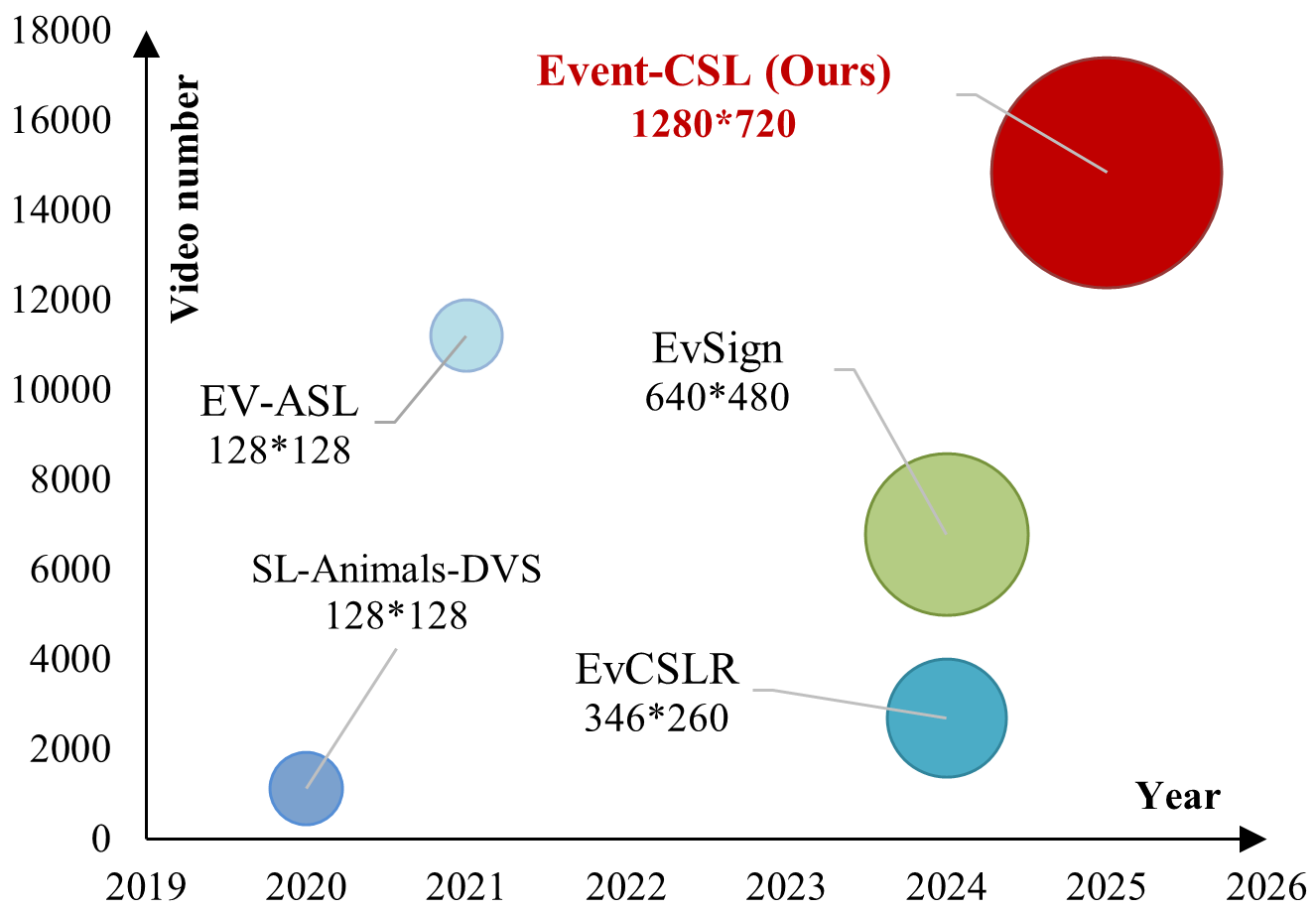}
\caption{Comparison between existing event-based sign language datasets and our newly proposed Event-CSL benchmark dataset. The bubble size represents the resolution.}  
\label{fig:firstIMG}
\end{figure}

% Although limited, there are already some research efforts that have begun to focus on event-based sign language translation, as shown in Fig.~\ref{fig:firstIMG}. Early models are typically developed based on simulated event data from existing RGB datasets, such as PHOENIX-2014~\cite{Koller2015Continuous} and CSL-Daily~\cite{zhou2021improving}. Obviously, the simulated data can't fully reflect the key features of the real event stream. Recently, three real event datasets have been proposed, including SL-Animal-DVS~\cite{vasudevan2020introduction}, EV-ASL~\cite{Wang2021Event-Based}, and EvSign~\cite{zhang2024evsign}. However, the first two datasets have limited spatial resolution ($128 \times 128$), and the EvSign~\cite{zhang2024evsign} crops the head regions of the actor for privacy-preserving. We believe this approach may corrupt the raw information of the event stream, significantly impacting the recognition and analysis of gestures involving the head area. 

Although research in this area remains limited, several studies have recently begun to explore event-based sign language translation (SLT) and sign language recognition (SLR), as illustrated in Fig.~\ref{fig:firstIMG}. Early approaches are typically developed using simulated event data generated from existing RGB datasets, such as PHOENIX-2014~\cite{Koller2015Continuous} and CSL-Daily~\cite{zhou2021improving}. However, such simulated data fail to fully capture the essential characteristics of real event streams.
Recently, four real event-based datasets have been introduced, namely SL-Animal-DVS~\cite{vasudevan2020introduction}, EV-ASL~\cite{Wang2021Event-Based}, EvCSLR~\cite{jiang2024evcslr}, and EvSign~\cite{zhang2024evsign}. The first three datasets are designed for the SLR task and suffer from low spatial resolutions ($128 \times 128$ and $346 \times 260$), while EvSign~\cite{zhang2024evsign} removes the actors’ head regions for privacy protection. We argue that this operation may distort the event signals, thereby negatively impacting the recognition and analysis of gestures involving head movements.
In addition, to reduce the substantial computational complexity introduced by spatial modeling, most existing SLT methods~\cite{Zhou_2023_ICCV, zhang2024evsign} primarily focus on temporal modeling, while overlooking the crucial role of spatial detail features in gesture understanding.

In this paper, we propose a new sign language translation dataset, termed Event-CSL, which is the largest, high-definition ($1280 \times 720$) real-event dataset collected using the Prophesee EVK4-HD event camera. It contains 14,827 videos, 14821 glosses, and 2,544 Chinese words in the text vocabulary. Our dataset is collected across diverse indoor and outdoor scenes, covering multiple viewpoints, lighting conditions, and camera motions. More importantly, we provide the most primitive event stream data, which lays the groundwork for accurately recognizing the meanings conveyed by gestures. In our experiments, we split these videos into training, validation, and testing subsets, which contain 12,602, 741, and 1,484 samples, respectively. To build a more comprehensive benchmark, we retrain and evaluate recently released SLT models for future work to compare. More details can be found in Section~\ref{eventSLT_dataset}.

% On the basis of our newly proposed Event-CSL dataset, we also propose a new event-based sign language translation framework, as shown in Fig.~\ref{framework}. The key insight of our framework is that Convolutional Neural Networks (CNNs) effectively capture local features and have consistently been the dominant visual backbone networks for sign language translation tasks. Despite the impact of Transformer models~\cite{vaswani2017Transformer}, their lower complexity and higher precision have continued to be favored. Therefore, this paper employs a residual network, ResNet18~\cite{He_2016_CVPR}, to encode the input event streams. To further enhance its global modeling capabilities, we introduce Mamba layers~\cite{Gu2023Mamba:} to fuse CNN feature maps along the temporal views, thereby strengthening its perception. This hybrid CNN-Mamba architecture can even achieve better sign language translation results than dense Transformer models. We input the event visual features into the decoder to translate and obtain the final Chinese descriptions. Extensive experiments conducted on multiple existing SLT datasets and our newly proposed Event-CSL all validated the effectiveness of our framework. 

Based on the newly proposed Event-CSL dataset, we further propose a novel event-based sign language translation framework, termed EvSLT. As shown in Fig.~\ref{framework}, we first feed the event video frames into a visual encoder, which is composed of residual networks~\cite{He_2016_CVPR}, to extract visual features. The resulting video features are then segmented into consecutive video clip features of a predefined length, following the temporal order of the frames. Features from these fixed-length video clip are subsequently compressed and aggregated using a Mamba-based memory aggregation module to capture fine-grained spatial details. Together with temporal features extracted via an average pooling layer and temporal convolution, the compressed clip features and aggregated spatial features are fed into a graph-guided spatiotemporal fusion module, where hypergraphs constructed from the compressed clip features serve as a bridge to enable effective interactions between spatial and temporal representations. Finally, the spatiotemporally fused features are projected into language model embedding space through a sign embedding layer, and then fed into a language model (mBART~\cite{Liu2020Multilingual}) for decoding into sign language text.

To sum up, the main contributions of this paper can be summarized as the following three aspects: 

1). We propose a new large-scale, high-definition event-based sign language translation dataset, termed Event-CSL. It contains 14,827 videos, 14,821 glosses, and 2,544 Chinese words recorded in both indoor and outdoor scenarios. 

2). We propose a novel event-based sign language translation framework, termed EvSLT. It employs a Mamba-based memory aggregation module to compress and aggregate spatial features, and introduces a graph-guided spatiotemporal fusion module for spatiotemporal interaction.

3). Extensive experiments conducted on two event-based SLT datasets, i.e., EvSign, and our Event-CSL, fully validated the effectiveness of our framework. 
% 3). Extensive experiments conducted on four event-based SLT datasets, i.e., EvSign,  PHOENIX-2014T-Event, CSL-Daily-Event, and Event-CSL, fully validated the effectiveness of our framework. 
% \begin{itemize}
%   \item  We propose a new large-scale, high-definition event-based sign language translation dataset, termed Event-CSL. It contains 14,827 videos and 2,544 Chinese words recorded in both indoor and outdoor scenarios. 
%   \item We propose a novel event-based sign language translation framework, termed EvSLT. It employs a Mamba-based memory aggregation module to compress and aggregate spatial features, and introduces a graph-guided spatiotemporal fusion module for spatiotemporal interaction.
%   \item Extensive experiments conducted on four event-based SLT datasets, i.e., EvSign,  PHOENIX-2014T-Event, CSL-Daily-Event, and Event-CSL, fully validated the effectiveness of our framework. 
% \end{itemize}
% In addition, we have evaluated multiple representative and state-of-the-art (SOTA) SLT algorithms on our newly proposed Event-CSL dataset and also two simulation datasets (PHOENIX-2014T-Event~\cite{camgoz2018neural}, CSL-Daily-Event~\cite{zhou2021improving}), hoping that these benchmarks can contribute to the advancement of the SLT field. 

\section{Related Work} 

\subsection{Sign Language Translation}
% In recent years, sign language translation has been widely studied in the field of computer vision. 
Sign language translation (SLT) aims to transform sign language videos into spoken text, thereby facilitating communication for deaf individuals. 
In earlier years, Camgöz et al.~\cite{camgoz2018neural} introduced the SLT task. 
% In the task of SLT, the word order and grammatical differences between sign language and spoken language are taken into account to convert sign language videos into spoken language translations. 
% Camgöz et al.~\cite{camgoz2020sign} propose an architecture based on Transformer, which can simultaneously learn sign language continuous recognition and translation end-to-end, without any ground-truth timing information, and improve performance. 
% Zhou et al.~\cite{Zhou_2023_ICCV} propose a two-stage approach, which first combines contrast language image pre-training (CLIP) with masking self-supervised learning to create a pre-task to bridge the semantic gap between visual and text representation and restore masking sentences. 
% Next, build an end-to-end architecture that inherits the parameters of the pre-trained visual encoder and decoder. 
% SignNet II~\cite{Chaudhary2023SignNet} enhances text-to-sign translation performance by jointly training sign-to-text and text-to-sign Transformer networks. 
Traditional gloss-based methods~\cite{camgoz2020sign, chen2022simple, ye2023cross} first translate sign language videos into glosses and then convert those glosses into spoken text. 
Joint‑SLT~\cite{camgoz2020sign} proposes a Transformer‑based framework that jointly learns continuous sign language recognition and translation in an end‑to‑end manner.
% Chen et al.~\cite{chen2022simple} propose a simple multimodal transfer learning baseline for gloss-based sign language translation.
Recently, gloss-free SLT methods~\cite{yin2023gloss, Lin2023Gloss-Free, Zhou_2023_ICCV, wong2024sign2gpt} have gained increasing attention for removing the need for costly gloss annotations. 
For example, 
% GASLT~\cite{yin2023gloss} does not rely on glosses and improves comprehension at the sentence level of sign language video by introducing knowledge transfer of natural language models. 
% GloFE~\cite{Lin2023Gloss-Free} extracts common concepts from text and presents them as indirect representations of weak forms, then uses global embeddings of these concepts as queries for cross-attention lookups to find the corresponding information in the learned visual features. 
Zhou et al.~\cite{Zhou_2023_ICCV} propose a two-stage approach, which bridges the semantic gap between visual and text representation and restores masking sentences. 
Sign2GPT~\cite{wong2024sign2gpt} leverages LLM for gloss-free sign language translation.
% Yao et al.~\cite{Yao2023Sign} continuously refine prototypes across attention mechanisms, iteratively optimize the semantic representation of sign language videos, imitate human reading behaviors, and finally generate fluent and accurate sign language translation. 
% Different from previous works, we exploit the event camera’s high sensitivity to motion to achieve more reliable sign language translation under challenging conditions such as fast movements, varying illumination, and both indoor and outdoor environments.
Different from previous works, we explore an event-based, gloss-free SLT task, leveraging the high dynamic range, high temporal resolution, and inherent privacy benefits of event cameras to enable accurate and reliable translation.

\subsection{State Space Model} 
State space models (SSMs)~\cite{kalman1960new} have recently gained attention due to their linear complexity. Gu et al.~\cite{Gu2021Efficiently} propose the Structured State Space sequence model, which reparameterizes the SSM, stabilizes the state matrix via low-rank correction, and simplifies computation through a Cauchy kernel formulation. Mamba~\cite{Gu2023Mamba:} is a sequence modeling approach with linear complexity that addresses the low computational efficiency of traditional methods on long sequences. In computer vision, VMamba~\cite{Liu2024VMamba:} extends the visual Mamba with four-directional scanning while retaining the benefits of a global receptive field and dynamic weights. Vim~\cite{Zhu2024Vision} further introduces a bidirectional vision Mamba block that serves as a universal backbone for vision tasks. Following the design of vision Mamba, many recent visual models~\cite{li2024videomamba, huang2024mamba, yang2024vivim, shi2025vmambair} adopt similar strategies to achieve efficient spatial modeling. Building upon these advances, this paper employs the vision Mamba block to aggregate spatial details from video clips, thereby effectively enhancing sign language translation performance.

\section{Methodology}  

% \begin{figure*}
% \centering
% \includegraphics[width=1\linewidth]{figures/EventSLT_framework2.jpg}
% \caption{\textbf{An overview of the proposed event-based Sign Language Translation framework, termed EvSLT}. The framework segments video frames into consecutive clips and employs a Mamba-based memory aggregation module to compress and aggregate spatial detail features. These spatial features and compressed clip features, together with the temporal representations generated by the temporal convolution, are then fed into a graph-guided spatiotemporal fusion module. In this module, hypergraphs are constructed from the compressed features to bridge the interaction between memory-enhanced spatial features and temporal representations. Finally, a Transformer-based language model, initialized with pre-trained weights from mBART~\cite{Liu2020Multilingual}, is employed to generate the corresponding Chinese sentences.}  
% \label{framework}
% \end{figure*}

\begin{figure*}
\centering
\includegraphics[width=1\linewidth]{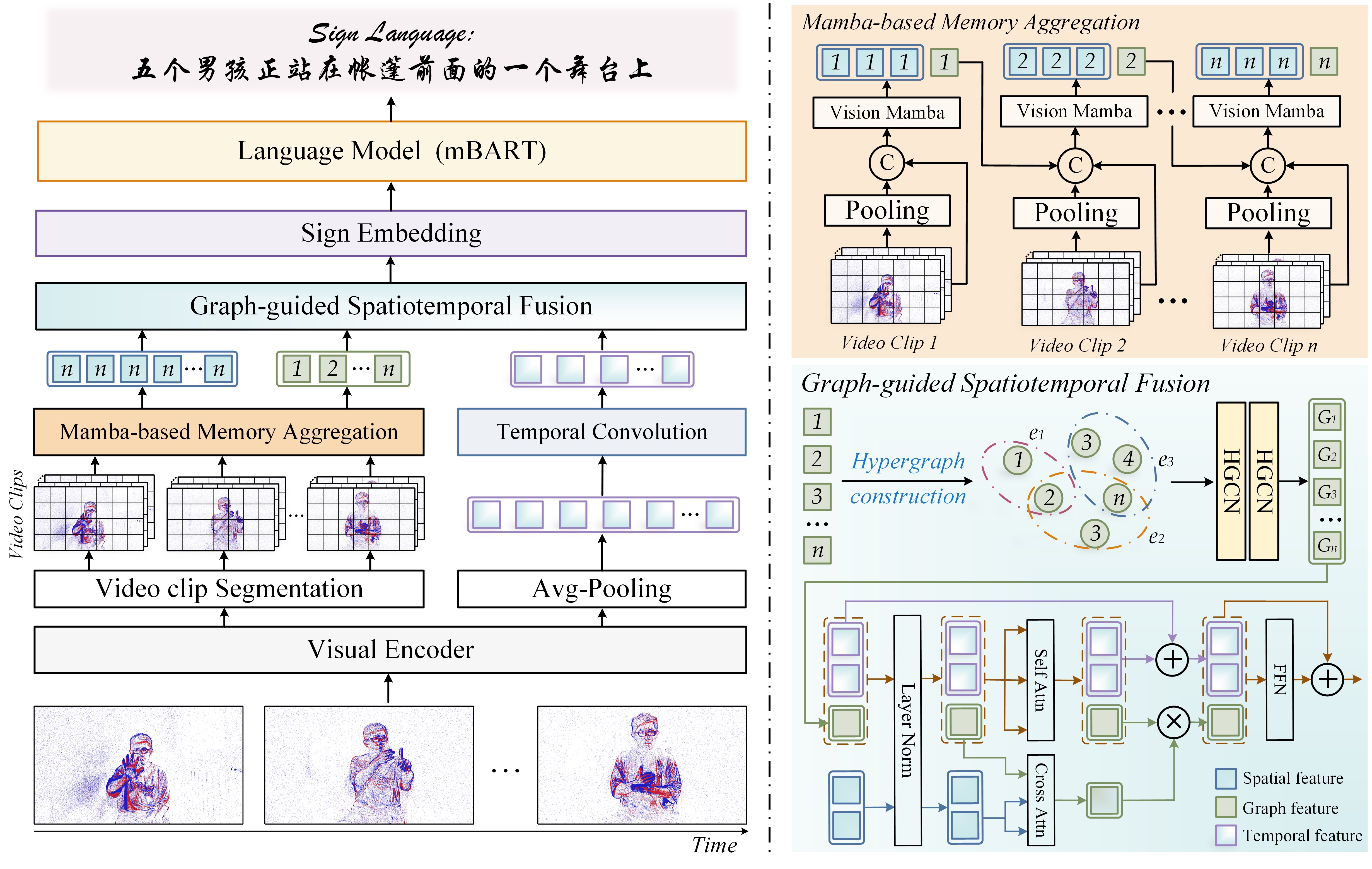}
\caption{\textbf{An overview of the proposed event-based sign language translation framework}. The framework segments continuous video features into clip features and employs a Mamba-based memory aggregation module to compress and aggregate spatial details. The resulting spatial and compressed clip features, together with temporal representations from temporal convolution, are fused in a graph-guided spatiotemporal fusion module. Hypergraphs built from compressed clip features facilitate interaction between memory-enhanced spatial and temporal representations. Finally, a Transformer-based language model generates the corresponding Chinese sentences.}  
\label{framework}
\end{figure*}

% \begin{figure*}
% \centering
% \includegraphics[width=1\linewidth]{figures/EventSLT_framework2.jpg}
% \caption{An overview of the proposed event-based sign language translation framework.}  
% \label{framework}
% \end{figure*}

 % In this section, we will first give an overview of our proposed event-based sign language translation framework. Then, we will introduce the details of input representation, network architecture, and loss function. 

\subsection{Overview}  
As illustrated in Fig.~\ref{framework}, we propose a unified framework for event-based sign language translation. Input event streams are first stacked into event frames and passed through a visual encoder to extract features. Temporal dynamics are modeled using average pooling followed by temporal convolution, while a Mamba-based memory aggregation module compresses and aggregates clip-level spatial representations. A graph-guided spatiotemporal fusion module then constructs relational graphs from the compressed clip features, enabling effective interaction between enhanced spatial details and temporal dependencies. Finally, the fused spatiotemporal features are projected via a sign embedding module and decoded by a language model (mBART~\cite{Liu2020Multilingual}) to generate accurate Chinese sentences.

\subsection{Input Representation}  
 Given event streams $\mathcal{E}^{p} = \{ e_{1}, e_{2}, ..., e_{M} \}$, where $M$ denotes the number of event points, each point $e_i|i \in (1, M)$ is a quadruple $\{x, y, t, p\}$, here the $(x, y)$ denotes the spatial coordinates, $t$ and $p$ denotes the time stamp and polarity, respectively. To make full use of existing deep neural networks, we first stack the event streams of each video into a sequence of event frames, denoted as $\mathcal{E}^{T} = \{ E_{1}, E_{2}, \ldots, E_{T} \}$, where $T$ denotes the number of stacked frames. Each event frame is then resized to a fixed resolution, i.e., $E_{i} \in \mathbb{R}^{C \times H \times W}$. Here, $\{C, H, W\}$ denotes the dimensions of the channel, height, and width of an event frame, respectively. During training, we randomly sample a mini-batch of \textit{B} event video samples as input, with dimensions $B \times T \times C \times H \times W$.

% Then, we feed the event frames into the CNN-Mamba Block for the local and global feature extraction. We adopt a stem network (a convolutional layer with filter $7 \times 7$ and max-pooling layer) to embed the event frames into their feature representations $F \in \mathbb{R}^{B \times T \times C \times H \times W}$. 

\subsection{Network Architecture}

\noindent $\bullet$ \textbf{Visual Encoder and Temporal Branch.}
Given an input of size $B \times T \times C \times H \times W$, we first employ a residual network as the visual encoder to extract visual feature representations for video frames. Specifically, we reshape the input by merging the batch and temporal dimensions into $B \times T \rightarrow B * T$, resulting in a tensor of size $B * T \times C \times H \times W$, and then fed into ResNet-18~\cite{He_2016_CVPR} to extract visual features, yielding video feature maps $\mathbf{F} \in \mathbb{R}^ {B \times T \times C' \times H' \times W'}$ ($H' = H / 32$, $W' = W / 32$). Here, $C'$, $H'$, and $W'$ denote the channel dimension, height, and width of the feature maps obtained after the hierarchical downsampling of ResNet-18. 

In the temporal branch, we perform global average pooling over the spatial dimensions of the extracted feature maps to obtain the temporal features $\mathbf{F}_{t} \in \mathbb{R}^ {B \times T \times C'}$. 
After that, we employ two temporal convolution blocks to aggregate temporal information and obtain the aggregated temporal features $\mathbf{F}_{t}' \in \mathbb{R}^ {B \times T' \times C'}$. It can be formulated as:
\begin{equation}
\begin{aligned}
    \mathbf{H}^{(1)} &= \text{MaxPool}\!\left(
        \sigma\!\left(\text{BN}\!\left(\text{Conv1d}_{5}^{(1)}\!\left(\mathbf{F}_{t}\right)\right)\right)
    \right), \\
    \mathbf{F}_{t}' &= \text{MaxPool}\!\left(
        \sigma\!\left(\text{BN}\!\left(\text{Conv1d}_{5}^{(2)}\!\left(\mathbf{H}^{(1)}\right)\right)\right)
    \right), \\
\end{aligned}
\end{equation}
where $\text{Conv1d}_{5}^{(i)}(\cdot)$ denotes the 1D convolution with kernel size 5 in the $i$-th temporal convolution block, $\text{MaxPool}$ represents the max pooling operation, and $\sigma(\cdot)$ denotes the ReLU activation function. 
% Through multi-layer temporal convolution, the model captures the intrinsic temporal dynamics of hand and body movements, enabling the establishment of coherent temporal dependencies. This temporal modeling is essential for enhancing contextual understanding, thereby supporting accurate and efficient sign language translation.
Temporal convolution enables the model to capture continuous hand movements and establish coherent temporal dependencies, thereby improving contextual understanding and translation accuracy.

\noindent $\bullet$ \textbf{Mamba-based Memory Aggregation.}
To preserve spatial detail while reducing computational cost, we introduce a Mamba-based memory aggregation module to compress and aggregate spatial features. Specifically, we first divide the continuous video features $\mathbf{F} \in \mathbb{R}^ {B \times T \times C' \times H' \times W'}$ into fixed-length video clip features along the temporal dimension, resulting in $n$ clip features, denoted as $\mathbf{F}_{c} = \{{\mathbf{F}_{c}^{1}, \mathbf{F}_{c}^{2}, \dots, \mathbf{F}_{c}^{n}}\}$. Consequently, the spatial feature of each video clip can be represented as $\mathbf{F}_{c}^{i} \in \mathbb{R}^{B \times \frac{T}{n} \times C' \times H' \times W'}$, which can be further reshaped into $\mathbf{F}_{c}^{i} \in \mathbb{R}^{B \times \frac{T \cdot H' \cdot W'}{n} \times C'}$. Next, we apply global average pooling to the spatial feature of the $i$-th video clip to compress them and obtain a compressed clip feature representation $\mathbf{F}_{g}^{i} \in \mathbb{R}^{B \times 1 \times C'}$. The resulting compressed clip feature is then concatenated with the spatial features of the corresponding video clip and fed into the visual Mamba network for spatial feature aggregation.

Following vision Mamba~\cite{Zhu2024Vision}, the concatenated vision features are first normalized through a normalization layer, and then projected into features \textit{x} and \textit{z} using two linear layers. 
Specifically, we also employ both forward and backward scanning to process the input. For each direction, we obtain $x'$ using \textit{the 1D convolution} layer and \textit{SiLU} activation function:
 \begin{equation}
    \label{silu_conv1d} 
    x' = SiLU(Conv({x})).
\end{equation} 
After that, the output $x'$ is projected to yield the input matrix $\textbf{B}\in \mathbb{R}^{N \times T}$, output matrix $\textbf{C}\in \mathbb{R}^{T \times N}$, and also the timescale parameter $\Delta$:
 \begin{equation}
    \label{linear} 
    \mathbf{B},\mathbf{C},\Delta = Linear(x'),
\end{equation} 
Here, the timescale parameter $\Delta$ is used for discretization, as the raw SSM is designed for a continuous system, but the data we process are discretized visual tokens. The Zero-Order Hold (ZOH) rule is adopted to discretize the continuous state space matrices:
% \begin{equation}
% h'=\overline{\mathbf{A}}h+\overline{\mathbf{B}}x',\quad y=\mathbf{\overline{C}}h',
% \end{equation}
% where h and $h'$ represent the discrete hidden state, $x'$ and y are the discrete inputs
% and outputs. $\overline{\mathbf{A}}$, $\overline{\mathbf{B}}$, and $\overline{\mathbf{C}}$ are discrete parameters of the system, and the process of discretization through time scale parameter $\Delta$ can be expressed as:

% \begin{equation}
% \begin{aligned}
% &\overline{\mathbf{A}}=\exp{(\Delta{\mathbf{A}})},\\
% &\overline{\mathbf{B}}=(\Delta{\mathbf{A}})^{-1}(\exp{(\Delta{\mathbf{A}})}-I)\cdot\Delta{\mathbf{B}},\\
% &\overline{\mathbf{C}} = {\mathbf{C}},
% % &\overline{\mathbf{A}},\overline{\mathbf{B}}=ZOH(\mathbf{A},\mathbf{B},\Delta). 
% \end{aligned}
% \end{equation}
\begin{equation}
    \overline{\mathbf{A}}, \overline{\mathbf{B}} = \text{ZOH}(\mathbf{A}, \mathbf{B}, \Delta),
    \label{eq:mamba_discretize}
\end{equation}
where ${\mathbf{A}}\in \mathbb{R}^{N \times N}$ is the state matrix. Therefore, the formula of the discretized SSM can be denoted as:
\begin{equation}
h'=\overline{\mathbf{A}}h+\overline{\mathbf{B}}x',\quad y=\mathbf{C}h',
\end{equation}
where $h$ and $h'$ represent the hidden state, $x'$ and $y$ are the inputs and output.
The forward and backward SSM branches are applied for bidirectional context awareness to produce $y_{for}$, $y_{back}$. After that, $y_{for}$, $y_{back}$ are gated by \textit{z} and the result of their addition is linearly projected to $y^{\prime}$, which can be written as:
\begin{equation}
\begin{aligned}
&y^{\prime}=Linear(y_{for}\odot SiLU(z)+y_{back}\odot SiLU(z)).
\end{aligned}
\end{equation}

For each video clip feature $\mathbf{F}_{c}^{i}$, we apply the aforementioned series of operations, and the resulting compressed clip features $\mathbf{F}_{g}^{(i)'}$ are pass to the $(i+1)$-th video clip feature to enable inter-clip information exchange and memory aggregation. Mathematically, this can be expressed as:
\begin{equation}
\begin{aligned}
\mathbf{F}_{g}^{(i+1)'}, \mathbf{F}_{c}^{(i+1)'} = Mamba([\mathbf{F}_{g}^{(i)'},\mathbf{F}_{g}^{i+1},\mathbf{F}_{c}^{i+1}]),
\end{aligned}
\end{equation}
where $[\cdot]$ denotes concatenation, and the compressed feature of the previous clip is discarded.

In this way, we progressively aggregate the detailed spatial features of each clip in a forward manner, analogous to how the human brain sequentially remember each video clip from beginning to end, resulting in the final aggregated spatial features $\mathbf{F_s} = \mathbf{F}_{c}^{(n)'}$ along with the compressed clip features $\mathbf{F}_{g} = [\mathbf{F}_{g}^{(1)'}, \mathbf{F}_{g}^{(2)'}, ..., \mathbf{F}_{g}^{(n)'}]$. Next, the memory-enhanced spatial features $\mathbf{F_s}$, the compressed clip features $\mathbf{F}_{g}$, and the temporal features $\mathbf{F}_{t}'$ are jointly fed into the graph-guided spatiotemporal fusion module.

\noindent $\bullet$ \textbf{Graph-guided Spatiotemporal Fusion.}
% The temporal features $\mathbf{F}_{t}'$ capture the sequential evolution of a sign language video from beginning to end, while the aggregated spatial features $\mathbf{F}_{s}$ preserve fine-grained spatial details. Meanwhile, the compressed clip features $\mathbf{F}_{g}$ provide the compact global representations of the video clips. 
To capture the complex relationships among different video clips, we treat the compressed feature vectors of video clips as nodes to construct hypergraphs. Following~\cite{chami2019hyperbolic}, we adopt K-nearest neighbors (K-NN, where the maximum number of neighboring nodes is set to 10 in our work) based on Euclidean distance to generate the hypergraph structure $\mathcal{H}$. It can be formulated as:

\begin{equation}
\begin{aligned}
\mathcal{H} &= \{V, \mathcal{E}, \mathbf{H}\}, \\[4pt] where \
\mathbf{H}_{j i} &= 
\begin{cases}
1, & \text{if } v_j \in e_i, \\
0, & \text{otherwise},
\end{cases} \\[6pt]
\end{aligned}
\end{equation}

\begin{equation}
\begin{aligned}
\mathbf{G} &= \mathbf{D}_v^{-\tfrac{1}{2}} \, \mathbf{H} \, \mathbf{D}_e^{-1} \, \mathbf{H}^\top \, \mathbf{D}_v^{-\tfrac{1}{2}},
\end{aligned}
\end{equation}
where $\mathbf{H}$ is the node–hyperedge incidence matrix, $V$ and $\mathcal{E}$ denote the vertex and hyperedge set. $\mathbf{D}_v$ and $\mathbf{D}_e$ are the node degree matrix and hyperedge degree matrix, respectively. $\mathbf{G}$ denotes the normalized hypergraph propagation matrix.
Subsequently, we feed $\mathbf{G}$ into Hypergraph Convolutional Network (HGCN)~\cite{chami2019hyperbolic} to facilitate effective interaction among video clips:

\begin{equation}
\begin{aligned}
\mathbf{X}' = \mathbf{G} (\mathbf{X} \mathbf{W} + \mathbf{b}),
\end{aligned}
\end{equation}
where the $\mathbf{X}$ and $\mathbf{X}'$ denote the input and output feature vectors, i.e., $\mathbf{F}_{g}$ and $\mathbf{F}_{g}'$, respectively. $\mathbf{W}$ is the learnable weight matrix and $\mathbf{b}$ is the bias.

To enable effective interaction between spatial and temporal features, i.e., $\mathbf{F}_{s}$ and $\mathbf{F}_{t}'$, we propose to use graph features $\mathbf{F}_{g}'$ derived from the compressed clip representations as a bridge to enable efficient spatiotemporal interaction. Specifically, we first pass the spatial, temporal, and graph features through a shared layer normalization layer. This is followed by two parallel attention networks that process the features concurrently. Let $SA$ and $CA$ denote the self-attention and cross-attention mechanisms, respectively. The temporal and graph features are concatenated and input to the $SA$ module, whereas the graph and spatial features serve as the query and context, respectively, for the $CA$ module. It can be expressed by the following formulas:
\begin{equation}
\begin{aligned}
\left[ \mathbf{F}_t^{\text{SA}}, \mathbf{F}_g^{\text{SA}} \right] &= \text{SA}\big([\mathbf{F}_{t}', \mathbf{F}_{g}']\big), \\
\mathbf{F}_g^{\text{CA}} &= \text{CA}\big( \mathbf{F}_{g}', \mathbf{F}_{s} \big),
\end{aligned}
\end{equation}
where the $\mathbf{F}_g^{\text{SA}}$ and $\mathbf{F}_g^{\text{CA}}$ denote the output graph features of $SA$ and $CA$, respectively. Subsequently, the two attention-enhanced graph features are combined via element-wise multiplication, together with the $SA$-enhanced temporal features, fed into a feedforward neural network. The feedforward network aggregates and transforms the attention-enhanced features, producing the final spatiotemporally enhanced fused representation.
Since the detailed spatial features consist of a large number of tokens, whereas the temporal and graph features contain relatively few, this design provides two key advantages. First, the graph features act as a bridge, facilitating effective interactions between the spatial and temporal features. Second, it substantially reduces the computational burden of the attention mechanism, since the spatial detail features are only used as the keys and values in the cross-attention operation.

\noindent $\bullet$ \textbf{Sign Language Decoder.} 
% Once we get the vision features $y^{\prime}$ from the aforementioned hybrid CNN-Mamba backbone, we adopt a temporal convolutional module to reduce the dimension along the view of the number of input frames (i.e., $T$). As shown in Fig.~\ref{framework}, it consists of a convolution layer, batch normalization (BN), ReLU, and max-pooling layers. The output will also be fed into a sign embedding module, which consists of a linear layer, BN, and a ReLU layer. Finally,  a language decoder is adapted to generate sign language sentences. In our practical implementation, we use the pre-trained parameters of \textit{mBART}~\cite{Liu2020Multilingual} to initialize the language model, which has been pre-trained on CC25~\cite{Liu2020Multilingual}, a multilingual corpus covering 25 languages. It consists of 3 Transformer encoder layers and 3 Transformer decoder layers. Note that we directly output the language based on given input event streams, without the help of gloss annotations, i.e., our model is a gloss-free sign language translation framework. 
Once the spatiotemporally enhanced features are obtained, they are fed into a sign embedding module, which consists of a linear layer and a ReLU activation. Finally, a language decoder is adapted to generate sign language sentences. In our practical implementation, we use the pre-trained parameters of \textit{mBART}~\cite{Liu2020Multilingual} to initialize the language model, which consists of 3 Transformer encoder layers and 3 Transformer decoder layers. Note that we directly output the language based on given input event streams, without the help of gloss annotations, i.e., our model is a gloss-free SLT framework.

\begin{table*}
\center
\scriptsize  
\caption{Comparison of the datasets for sign language recognition and translation.} 
\label{datasetlist} 
% \resizebox{\textwidth}{!}{ 
\begin{tabular}{l|cccccccccccccc}
\hline \toprule [0.5 pt]
\textbf{Dataset}    &\textbf{Year}  &\textbf{Language}   &\textbf{\#Videos}   &\textbf{Resolution}  &\textbf{Gloss}  &\textbf{Text}  &\textbf{Source} &\textbf{Raw Data} &\textbf{Scene} 	&\textbf{Continuous}  &\textbf{SLT} &\textbf{Event}\\ 
\hline 
\textbf{SIGNUM}~\cite{von2010signum}            &2010 	&DGS   &$15,075$   &$776\times578$   &$455$   &\-     &Lab  &-  &- 	&\cmark 	&\xmark		&\xmark 	\\ 
\textbf{DEVISIGN-G}~\cite{chai2015devisign}     &2015 	&CSL   &$432$ 	&$640\times480$  &$36$	&-     &Lab   &-  &- 	&\xmark &\xmark	 &\xmark \\ 
\textbf{DEVISIGN-D}~\cite{chai2015devisign}     &2015		&CSL   &$6,000$ 	&$640\times480$ 	&$500$	&-  	   &Lab  &-  &- 	 &\xmark &\xmark	 &\xmark 	\\ 
\textbf{DEVISIGN-L}~\cite{chai2015devisign}     &2015 		&CSL   &$24,000$ 		&$640\times480$ 	&$2,000$	&-  	   &Lab   &-  &-  	&\xmark 		&\xmark	 &\xmark 	\\ 
\textbf{PHOENIX-2014}~\cite{Koller2015Continuous}  	&2015  	&DGS    &$6,841$ 	&$210\times260$	 &$1,081$ 	&-       &TV  &-  &-  	&\cmark 	&\xmark 	&\xmark \\ 
\textbf{PHOENIX-2014T}~\cite{camgoz2018neural}  &2018 	&DGS    &$8,257$  &$210\times260$  	&$1,066$  	 &$2,887$      &TV  &-  &- 	&\cmark &\cmark		&\xmark  \\ 
\textbf{MSASL}~\cite{Joze2018MS-ASL:}   &2018 	&ASL     &$25,513$ 		&- 		&$1,000$ 	&-      &Web  &-  &-		&\xmark 		&\xmark		&\xmark \\ 
\textbf{INCLUDE}~\cite{Sridhar2020INCLUDE:}  &2020 	&ISL     &$4,287$ 	&$1920\times1080$ 		&$263$ 	&-  	   &Lab  &-  &- 	&\xmark 	&\xmark		&\xmark \\ 
\textbf{CSL-Daily}~\cite{zhou2021improving} &2021 	&CSL    &$20,654$   &$1920\times1080$   &$2,000$  &$2,343$      &Lab	 &-  &-  &\cmark 	&\cmark		&\xmark \\
% \textbf{How2Sign}~\cite{duarte2021how2sign} &2021 	&ASL    &   &-   &-  &-      &Lab	 &-  &-  &\cmark 	&\cmark		&\xmark \\
\hline 
\textbf{SL-Animals-DVS}~\cite{vasudevan2020introduction}  &2020	 &SSL   &$1,121$  &$128\times128$	&$19$ 	&-      &Lab  &\cmark  &Indoor  &\xmark	&\xmark		&\cmark	 	\\ 
\textbf{EV-ASL}~\cite{Wang2021Event-Based}  &2021	&ASL    &$11,200$ 		&$128\times128$		&$56$ &-  	    &Lab   &-  &-  &\xmark 		&\xmark 	&\cmark  	\\ 
\textbf{EvCSLR}~\cite{jiang2024evcslr}  &2024   &CSL   &$2,685$   &$346\times260$	&423  &- &Lab   &\cmark  &Indoor 	&\cmark  &\xmark		&\cmark	\\
\textbf{EvSign}~\cite{zhang2024evsign} &2024 &CSL     &$6,773$ 	&$640\times480$		&$1,387$ 	&$1,947$     &Lab	 &\cmark  &Indoor 	&\cmark  &\cmark		&\cmark	\\ 
\hline 
\textbf{Event-CSL (Ours)}  &2025  &CSL    &$\textbf{14,827}$  &$\mathbf{1280\times720}$	&$\textbf{14,821}$ 	&$\textbf{2,544}$     &Lab	 &\cmark  &\makecell[c]{Indoor, \\ Outdoor}   &\cmark  &\cmark	&\cmark  \\ 
\hline \toprule [0.5 pt]
\end{tabular}
% } 
\end{table*}	

\subsection{Loss Function} 
% The language decoder sequentially generates the logits of sign language sentences $\mathbf{S}=\{p(<BOS>),p(w_{1}),p(w_{2}),...,p(w_{U}),p(<EOS>)\}$. In sentence $\mathbf{S}$, the first special word $<BOS>$ marks the beginning of the sentence, the last special word $<EOS>$ marks the end of the sentence, and $p(w)$ represents the logits of the generated word. In this work, we adopt the widely used \textit{cross-entropy loss} function to calculate the distance between the annotated sign language and the generated sentence, which can be formulated as: 
% \begin{equation}
%     \mathcal{L}=-\log \prod_{u=1}^Up(s_u|w_u). 
% \end{equation} 

The language decoder sequentially generates the logits of sign language sentence S, which can be expressed as:
% \begin{equation}
% $\mathbf{S}=\{p({<}bos{>}),p(w_{1}),p(w_{2}),...,p(w_{U}),p({<}eos{>})\}$
% \end{equation
\begin{equation}
    S=\{p({<}bos{>}),p(w_1),p(w_2),...,p(w_U),p({<}eos{>})\}.
\end{equation}
The first special word $<$bos$>$ marks the beginning of the sentence, the last special word $<$eos$>$ marks the end of the sentence, and $p(w)$ represents the logits of the generated words. In this work, we adopt the widely used \textit{cross-entropy loss} function to calculate the distance between the annotated sign language sentence and the generated sentence, which can be formulated as: 
\begin{equation}
    \mathcal{L}=-\log \prod_{u=1}^Up(s_u|w_u). 
\end{equation}

\section{Event-CSL Dataset} \label{eventSLT_dataset}
% In this section, we introduce the details of the proposed sign language dataset, Event-CSL. Firstly, when collecting sign language videos, we conduct the recording in accordance with the protocols. Next, we perform Statistical Analysis on the collected sign language videos and annotations. Finally, we consider some sign baselines to construct Benchmark Baselines.

% \begin{figure*}[!htp]
% \centering
% \includegraphics[width=7in]{figures/EventSLT_demos.jpg}
% \caption{Representative samples from our proposed Event-CSL dataset and the corresponding text annotations} 
% \label{fig:images10}
% \end{figure*}

\subsection{Protocols} 
When collecting our sign language translation dataset Event-CSL, we follow the following protocols: 
\text{\textit{1). Diversity of views}:} We capture different perspectives of sign language, including slightly downward, horizontal, and slightly upward views.
\text{\textit{2). Camera motions}:} 
% The movement of the event camera significantly affects the number of events collected and the complexity of the background. 
Our dataset collection takes into account a variety of motion patterns, including fast movement, slow movement, and moderate movement.  
\text{\textit{3). Diversity of scenes}:} In order to conform to the daily use of sign language, we consider different usage scenarios when recording videos. Specifically, these scenarios comprise office environments, daily life, and outdoor scenes (such as streets, parking lots, and open spaces).
\text{\textit{4). Intensity of light}:} In environments with strong or weak light intensity (high and low exposure), event cameras can still record information about the movement of objects due to the presence of pixel changes. We record videos under different lighting conditions, including strong, low, and dim light. 
\text{\textit{5). Continuity of action}:} Keep the movement of sign language flowing and coherent when recording sign language, reducing useless information and redundant video frames. 
\text{\textit{6). Diversity of sign language content}:} To better reflect real-world applications, sign language videos should cover diverse content. We randomly extract captions from the COCO-CN~\cite{tmm2019-cococn} dataset, encompassing daily life, outdoor activities, animals and plants, sports, character introductions, and landscapes.

\subsection{Statistical Analysis} 
\begin{figure*}
\centering
\includegraphics[width=6.5in]{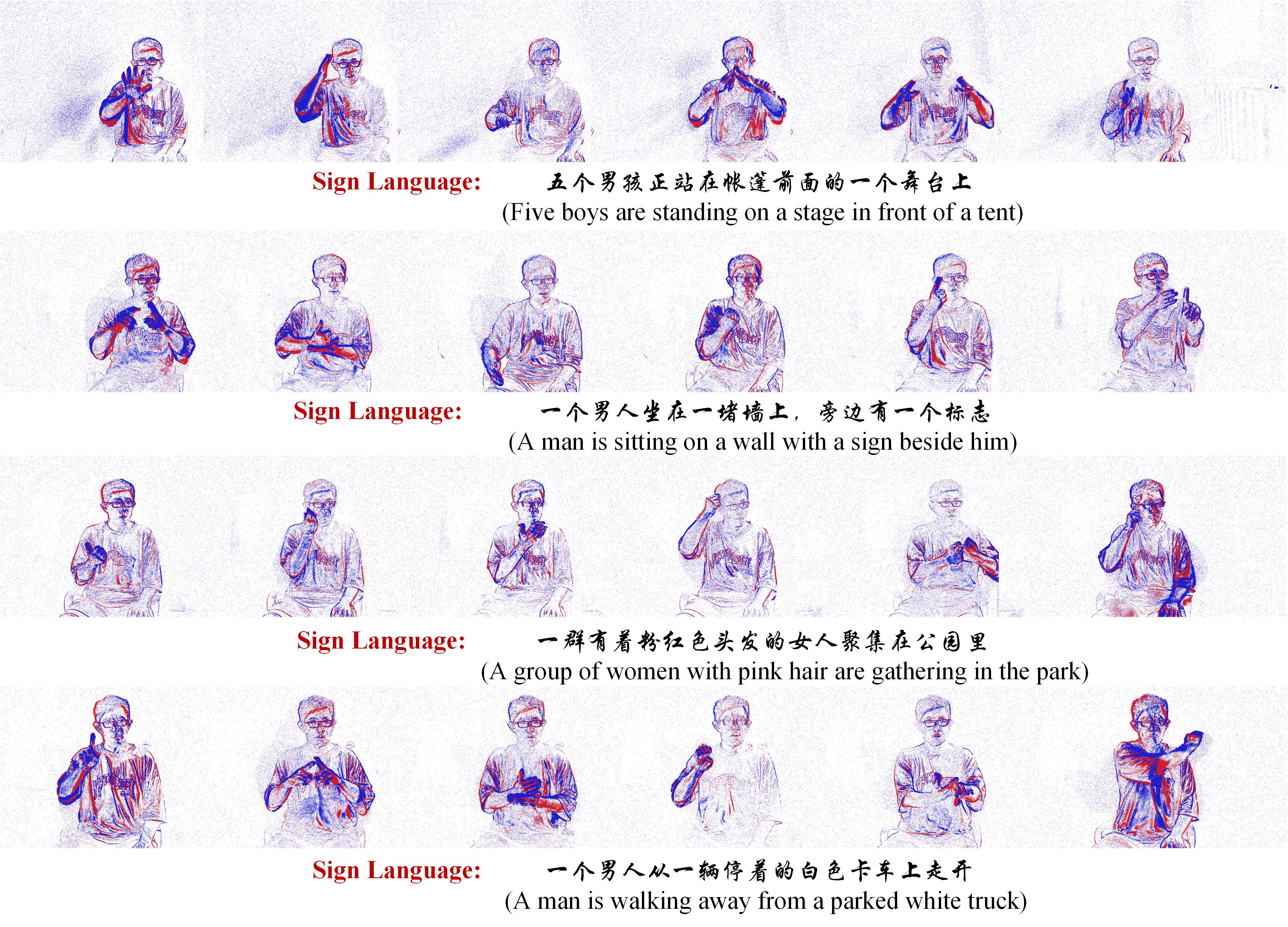}
\caption{Representative samples from our proposed Event-CSL dataset and the corresponding text annotations} 
\label{fig:images10}
\end{figure*}

\begin{figure}
    \centering
    \includegraphics[width=1\linewidth]{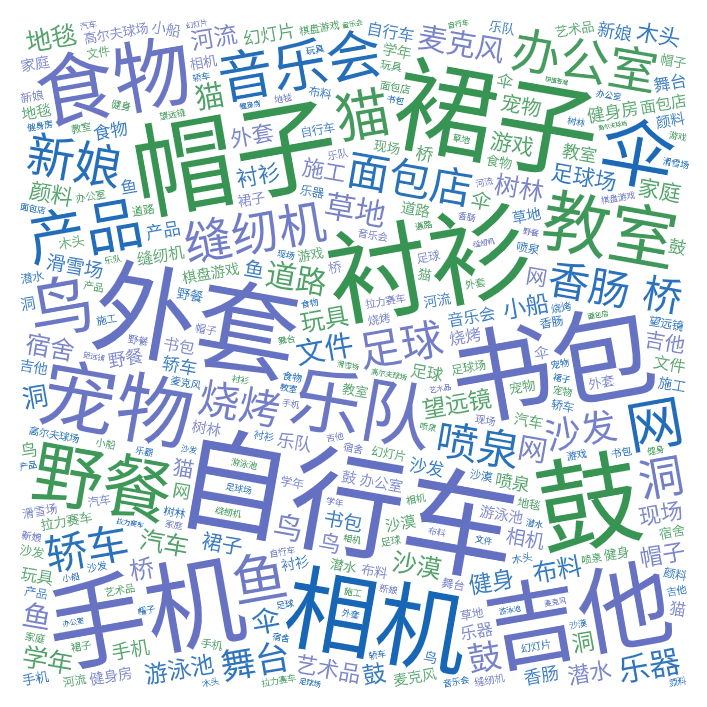}
    \caption{The word cloud visualization for the textual annotations in our proposed Event-CSL dataset.}
    \label{fig:wordCloud}
\end{figure}

% Our proposed dataset is the first large-scale, high-definition event-based sign language benchmark dataset. All the videos in our dataset are collected by the Prophesee EVK4-HD event camera, which has a resolution of $1280\times720$. Our dataset collects a total of 14,827 sign language videos and contains 14,821 glosses as well as 2,544 Chinese words in the text vocabulary. We randomly split the dataset into three subsets for training, validation, and testing, with their quantities being 12,602, 741, and 1,484, respectively. Additionally, within the text sentences of the dataset, the minimum, maximum, and average numbers of characters of the sentences are 8, 75, and 1,8, respectively. As shown in \ref{datasetlist}, the dataset proposed by us surpasses the existing event-based sign language datasets in terms of the number of videos, video resolution, gloss, and text vocabulary. 
% The visualization of the word cloud is provided in Fig.~\ref{fig:wordCloud}. 
Our proposed dataset is the first large-scale, high-definition event-based sign language benchmark. All videos are recorded using the Prophesee EVK4-HD event camera at a resolution of $1280\times720$. In total, the dataset contains 14,827 sign language videos, 14,821 gloss annotations, and a vocabulary of 2,544 Chinese words.
We randomly divide the dataset into training, validation, and test sets with 12,602, 741, and 1,484 videos, respectively. The text corpus includes sentences ranging from 8 to 75 characters, with an average length of 18.
As shown in Table~\ref{datasetlist}, our dataset significantly surpasses existing event-based sign language datasets in terms of video quantity, spatial resolution, gloss annotations, and vocabulary diversity.
The representative samples from Event-CSL with their corresponding text annotations are illustrated in Fig.~\ref{fig:images10}, and the word-cloud visualization is shown in Fig.~\ref{fig:wordCloud}.

\begin{figure*}[!htp]
\centering
\includegraphics[width=6.5in]{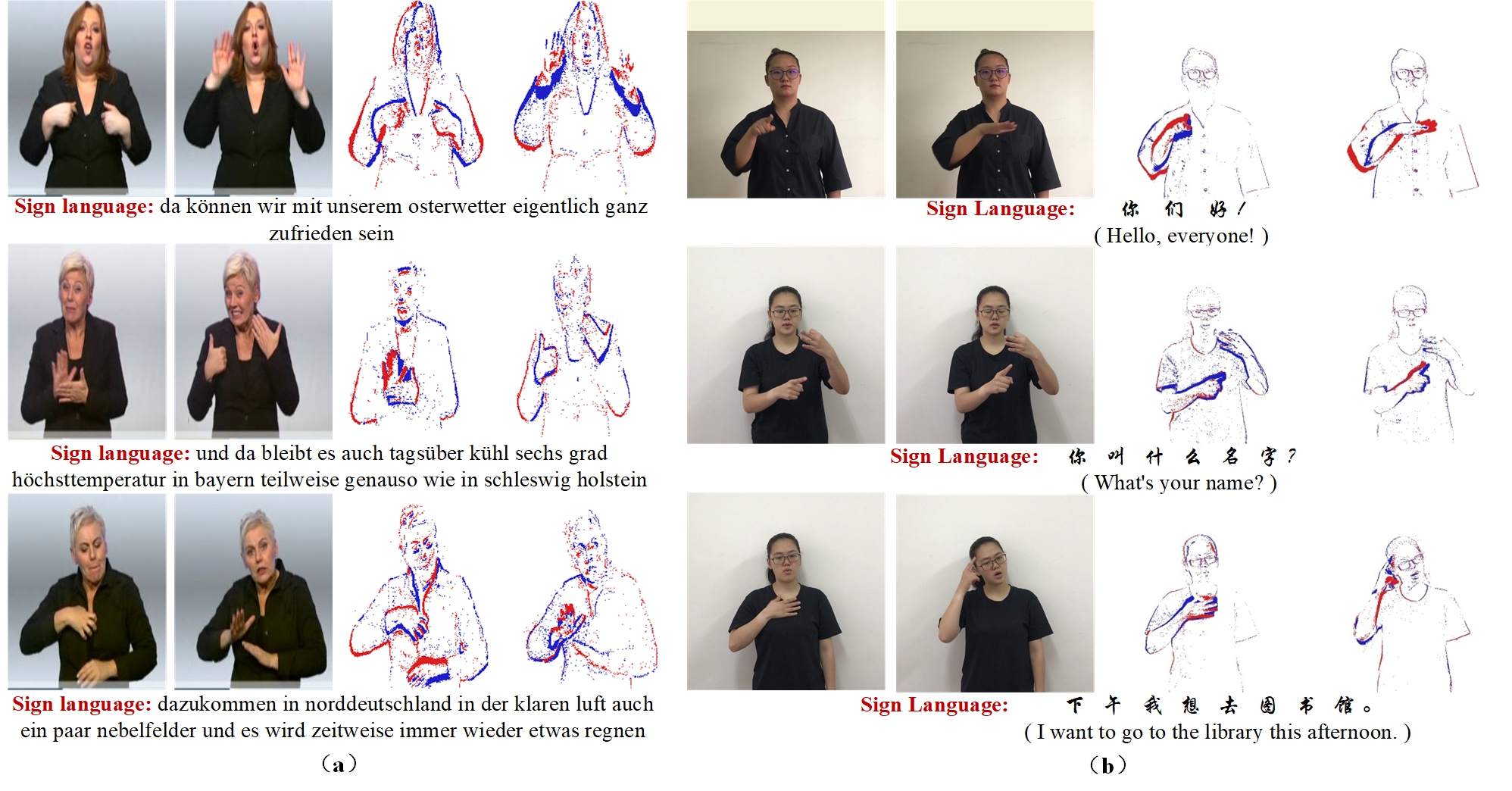}
\caption{The samples of video frames on the PHOENIX-2014T and CSL-Daily simulation datasets} 
\label{fig:simulation}
\end{figure*}

% \begin{figure}
%     \centering
%     \includegraphics[width=1\linewidth]{figures/word.png}
%     \caption{The word cloud visualization for the textual annotations in our proposed Event-CSL dataset.}
%     \label{fig:wordCloud}
% \end{figure}

\begin{table*}
\center
\small   
\caption{Experimental results on our Event-CSL dataset.}  
\label{SL-Event_result}
% \resizebox{\columnwidth}{!}{ 
\begin{tabular}{c|c|c|c|c|c|c|c|c}
\hline \toprule [0.5 pt] 
\textbf{No.} & \textbf{Algorithm} & \textbf{Publish}  & \textbf{Backbone} &\textbf{ROUGE}  &\textbf{BLEU-1} &\textbf{BLEU-2} &\textbf{BLEU-3} &\textbf{BLEU-4} \\
\hline
% \rowcolor[HTML]{F2F8FC}
% \multicolumn{9}{c}{Gloss-based} \\
01 &Joint-SLT~\cite{camgoz2020sign}     &CVPR2020   &ViT  &27.76	&30.30	&19.15	&12.63	&8.99\\ 
02 &Chen et al.~\cite{chen2022simple}     &CVPR2022   &S3D  &23.15	&24.35	&13.93	&8.01	&5.11\\ 
03 &Sign-XmDA~\cite{ye2023cross}     &ArXiv2023   &ViT  &31.48	&34.60	&23.56	&16.62	&12.39 \\
\hline
04 &GASLT~\cite{yin2023gloss}    &CVPR2023  &ViT  &31.69	&34.27	&23.41	&16.73	&12.64 \\
% \rowcolor[HTML]{F2F8FC}
% \multicolumn{9}{c}{Gloss-free} \\
05 &TSPNet~\cite{li2020tspnet}     &NeurIPS2020   &I3D   &26.63	&30.91	&17.84	&10.89	&7.25\\ 
06 &GFSLT~\cite{Zhou_2023_ICCV}    &ICCV2023  &CNN  &67.23 	&69.00 	&60.62 	&53.77 	&48.20 \\ 
07 &SignCL~\cite{ye2024improving}    &NeurIPS2024  &CNN  &67.76 &69.12 	&60.88 	&54.18 	&48.52	 \\ 
08 &Wang et al.~\cite{wang2024event}    &ArXiv2024  &CNN, Mamba  &68.02  &69.71 	&61.35	&54.40 	&48.70   \\ 
09 &Sign2GPT~\cite{wong2024sign2gpt}    &ICLR2024  &ViT &68.74  &70.35 	&61.98 	&55.13 	&49.23	 \\
% 10 &MMSLT~\cite{kim2025leveraging}    &ICCV2025  &CNN  &69.64 	&71.67 	&63.03 	&56.17	&50.65 \\ 
% 10 &MLSLT~\cite{tan2025multilingual}    &ACL2025  &  & 	& 	& 	&	& \\ 
% 11 &Jang et al.~\cite{jang2025lost}    &CVPR2025  &  & 	& 	& 	&	& \\ 
% 07 &Ours (\textit{w/o} Mamba)    &-  &CNN  &67.72  &69.21	&60.94	&54.20 	&48.68  \\ 
\hline 
10 &Ours    &-  &CNN  &\textbf{68.93}  &\textbf{70.40} &\textbf{62.09}	&\textbf{55.34} 	 	&\textbf{49.80}   \\ 

% \hline
% \#05 &\textbf{GFSLT-VLP}~\cite{Zhou_2023_ICCV}    &ICCV2023  &CNN+Transformer  &	   & 	& 	& 	&   \\ 
\hline \toprule [0.5 pt]
\end{tabular}
% }
\end{table*}

\begin{table*}
\center
\small   
\caption{Comparison between our model and other SOTA algorithms on the EvSign dataset.}
\label{EvSign_result}
\resizebox{\textwidth}{!}{ 
\begin{tabular}{c|l|c|c|c|c|c|c|c|c}
\hline \toprule [0.5 pt] 
\textbf{No.} & \textbf{Algorithm} &\textbf{Publish} &\textbf{Backbone} & \textbf{Modality}  &\textbf{ROUGE}  &\textbf{BLEU-1} &\textbf{BLEU-2} &\textbf{BLEU-3} &\textbf{BLEU-4} \\
\hline
01 &Joint-SLT.~\cite{camgoz2020sign}   &CVPR2020 &ViT  &RGB     &40.05	&39.84	&23.54	&15.60	&10.63 \\ 
02 &VAC+TH~\cite{min2021visual}   &ICCV2021  &CNN,BiLSTM    &RGB  &39.08	&38.74	&23.90	&15.88	&10.19 \\ 
03 &CorrNet+TH~\cite{hu2023continuous}  &CVPR2023 &CNN,BiLSTM    &RGB  &39.41	&39.45	&23.74	&15.68	&10.57 \\ 
\hline
04 &Joint-SLT~\cite{camgoz2020sign}  &CVPR2020 &ViT    &Event  &41.54	&40.13	&24.36	&16.04	&10.87 \\ 
05 &VAC+TH~\cite{min2021visual} &ICCV2021  &CNN,BiLSTM   &Event   &39.48	&39.22	&24.11	&15.94	&10.01 \\
06 &CorrNet+TH~\cite{hu2023continuous}  &CVPR2023 &CNN,BiLSTM   &Event  &41.23 &40.85	&25.34 	&16.95 	&11.83 \\
07 &GFSLT~\cite{Zhou_2023_ICCV}    &ICCV2023  &CNN  &Event 	&51.98 	&54.26 	&35.91 	&24.97 &17.88 \\ 
08 &Zhang et al.~\cite{zhang2024evsign}  &ECCV2024  &CNN,BiLSTM,ViT    &Event  &42.43 &41.44	&25.61 	&17.55 	&12.37 \\ 
% 08 &Wang et al.~\cite{wang2024event}  &ArXiv2024 &CNN,Mamba   &Event   &61.32  &61.31 	&46.76 	&36.68 	&29.41   \\ 
\hline
% 09 &Ours  &- &CNN   &Event   &\textbf{58.21}  &\textbf{58.02} 	&\textbf{53.02} 	&\textbf{48.88} 	&\textbf{45.46}   \\ 
09 &Ours  &- &CNN   &Event   &\textbf{53.43}  &\textbf{55.45} 	&\textbf{38.45} 	&\textbf{27.41} 	&\textbf{20.20}   \\ 
% \hline
% \#05 &\textbf{GFSLT-VLP}~\cite{Zhou_2023_ICCV}    &ICCV2023  &CNN+Transformer  &	   & 	& 	& 	&   \\ 
\hline \toprule [0.5 pt]
\end{tabular}
}
\end{table*}

\subsection{Benchmark Baselines}  
To construct a comprehensive benchmark dataset for event-based sign language translation, we retrain and report the performance of the following sign language translation algorithms as the baselines on the Event-CSL dataset for future works to compare:  
\textbf{1). Gloss-based SLT:}  Joint-SLT~\cite{camgoz2020sign}, Chen et al.~\cite{chen2022simple}, Sign-XmDA~\cite{ye2023cross}; 
\textbf{2). Gloss-free SLT:} TSPNet~\cite{li2020tspnet}, GASLT~\cite{yin2023gloss}, GFSLT~\cite{Zhou_2023_ICCV}, SignCL~\cite{ye2024improving}, Sign2GPT~\cite{wong2024sign2gpt}, and Wang et al.~\cite{wang2024event}.

\section{Experiment} 

% \begin{figure*}[!htp]
% \centering
% \includegraphics[width=7in]{figures/event_simulation.jpg}
% \caption{The samples of video frames on the PHOENIX-2014T and CSL-Daily simulation datasets} 
% \label{fig:simulation}
% \end{figure*}

\begin{table*}
\center
\small    
\caption{Comparison between our model and other SOTA algorithms on the PHOENIX-2014T-Event dataset.}
\label{PHOENIX-2014T-Event_result}
% \resizebox{\textwidth}{!}{ 
\begin{tabular}{c|l|c|c|c|c|c|c|c}
\hline \toprule [0.5 pt]
\textbf{No.} & \textbf{Algorithm} & \textbf{Publish}  & \textbf{Backbone} &\textbf{ROUGE}  &\textbf{BLEU-1} &\textbf{BLEU-2} &\textbf{BLEU-3} &\textbf{BLEU-4} \\
\hline
01 &TSPNet~\cite{li2020tspnet}     &NeurIPS2020   &I3D  &18.16				
 &19.41 &8.61 &5.92 &4.67  \\ 

02 &Joint-SLT~\cite{camgoz2020sign}     &CVPR2020  &ViT  &20.80 &21.25 &13.60 &9.96 &7.89 \\ 

03 &Sign-XmDA~\cite{ye2023cross}     &ArXiv2023   &ViT  &24.21 &24.80 &16.24 &11.91 &9.36   \\ 

04 &GASLT~\cite{yin2023gloss}    &CVPR2023  &ViT  &24.94 	&25.81 	&17.00 	&12.47 	&9.82   \\ 
 
05 &GFSLT~\cite{Zhou_2023_ICCV}    &ICCV2023  &CNN &33.45 &34.66	&24.63 	&18.49 	&14.72 	   \\ 
\hline
06 &Ours &-  &CNN  &\textbf{34.98 }	&\textbf{35.60} &\textbf{25.98} &\textbf{20.18}  &\textbf{16.48}   \\ 
\hline \toprule [0.5 pt] 
\end{tabular}
% }
\end{table*}

\begin{table*}
\center
\small   
\caption{Comparison between our model and other SOTA algorithms on the CSL-Daily-Event dataset.}
\label{CSL-Daily-Event_result}
% \resizebox{\columnwidth}{!}{ 
\begin{tabular}{c|l|c|c|c|c|c|c|c}
\hline  \toprule [0.5 pt] 
\textbf{No.} & \textbf{Algorithm} & \textbf{Publish}  & \textbf{Backbone} &\textbf{ROUGE}  &\textbf{BLEU-1} &\textbf{BLEU-2} &\textbf{BLEU-3} &\textbf{BLEU-4} \\
\hline
01 &TSPNet~\cite{li2020tspnet}     &NIPS2020   &I3D  &10.01	&10.69	&4.43	&2.36	&1.45
 \\

02 &Joint-SLT~\cite{camgoz2020sign}     &CVPR2020   &ViT  &12.68 	&13.47 	&6.63 	&3.79 	&2.44 \\ 

03 &Sign-XmDA~\cite{ye2023cross}     &ArXiv2023   &ViT  &13.84 	&14.50 	&7.75 	&4.50 	&2.87    \\ 

04 &GASLT~\cite{yin2023gloss}    &CVPR2023  &ViT  &15.25  &14.42  &8.48 	&5.20  &3.39 \\ 

05 &GFSLT~\cite{Zhou_2023_ICCV}    &ICCV2023  &CNN  &20.81  &20.82	   &13.42 	&8.98 	&6.37   \\ 
% 05 &GFSLT~\cite{Zhou_2023_ICCV}    &ICCV2023  &CNN  &	   &	& 	& 	&   \\ 
% 06 &Wang et al.~\cite{wang2024event}    &ArXiv2024  &CNN,Mamba  &24.89	   &25.69 	&16.94 	&11.54 &8.27   \\ 
\hline
06 &Ours    &-  &CNN  &\textbf{27.06}  &\textbf{27.14} 	&\textbf{15.68} 	&\textbf{10.13} &\textbf{7.09}   \\ 
\hline \toprule [0.5 pt] 
\end{tabular}
% }
\end{table*}

\subsection{Dataset and Evaluation Metric}  

Our experiments are conducted on four event-based sign language translation datasets, including the newly proposed \textbf{Event-CSL}, and public \textbf{EvSign}~\cite{zhang2024evsign},  \textbf{PHOENIX-2014T-Event}~\cite{camgoz2018neural}, \textbf{CSL-Daily-Event}~\cite{zhou2021improving} datasets.  

\noindent $\bullet$ \textbf{EvSign Dataset}~\cite{zhang2024evsign} is an event-based sign language dataset collected from daily-life scenarios, including social interactions, education, shopping, travel, and healthcare. It is recorded by multiple sign language volunteers, and its annotations are extracted and reorganized from the National Sign Language Dictionary of China and CSL-Daily to form fluent oral sentences. The training, validation, and testing sets contain 5,570, 553, and 650 sign language videos, respectively. In total, EvSign includes 1,387 Chinese glosses and a vocabulary of 1,947 Chinese words. 

\noindent $\bullet$ \textbf{PHOENIX-2014T-Event Dataset}~\cite{camgoz2018neural} is derived from German Sign Language weather broadcast videos. The original PHOENIX-2014T dataset consists of RGB video frames captured by conventional cameras. To generate an event-based version, we use DVS-Voltmeter~\cite{Lin2022DVS-Voltmeter:}, a stochastic process–based event simulator for dynamic vision sensors, as illustrated in Fig.~\ref{fig:simulation}(a). We denote the resulting event representation as PHOENIX-2014T-Event, which preserves the same dataset configuration as PHOENIX-2014T in terms of the number of videos and German annotations. The training, validation, and testing sets contain 7,096, 519, and 642 sign language videos, respectively. In total, the PHOENIX-2014T-Event dataset includes 1,066 German glosses and a vocabulary of 2,887 German words. 

\begin{table*}
\center
\small     
\caption{Component analysis results on the Event-CSL dataset.} 
\label{CAResults} 
% \resizebox{\columnwidth}{!}{
\begin{tabular}{c|cccc|c|c|c|c|c|c} 		
\hline 
\textbf{No.} & \textbf{Temporal}  &\textbf{Spatial} &\textbf{Mamba}  &\textbf{GSTF} &\textbf{ROUGE}  &\textbf{BLEU-1} &\textbf{BLEU-2} &\textbf{BLEU-3} &\textbf{BLEU-4}  &\textbf{Params} \\
\hline 
1 &\cmark   &\xmark    &\xmark   &\xmark   &67.23 	&69.00 	&60.62 	&53.77 	&48.20   & 115.5M \\
2 &\xmark   &\cmark    &\cmark   &\xmark   &56.31 	 &57.15 	  &47.77 	&40.63 	 & 35.12     & 119.8M  \\
\hline
3 &\cmark   &\cmark    &\xmark   &\cmark         &67.69 	&69.24 	&69.98 	&54.21 	&49.20      & 134.4M  \\
4 &\cmark   &\cmark    &\cmark   &\xmark   &67.65 	&69.27 	&60.76 	&53.98 	  &48.84    & 130.7M \\
5 &\cmark   &\cmark    &\cmark   &\cmark   &\textbf{68.93}  &\textbf{70.40} &\textbf{62.09}	&\textbf{55.34} 	 	&\textbf{49.80}   & 138.2M     \\

\hline
\end{tabular}
% }
\end{table*}

\noindent $\bullet$ \textbf{CSL-Daily-Event Dataset}~\cite{zhou2021improving} is derived from daily life scenarios, covering multiple themes such as family life, healthcare, and school life. We also obtain its event representation through the DVS-Voltmeter~\cite{Lin2022DVS-Voltmeter:}, as shown in Fig.\ref{fig:simulation} (b). We refer to the event representation of CSL-Daily as CSL-Daily-Event, and it remains consistent with the dataset configuration of CSL-Daily in terms of the number of videos and Chinese annotations. The training set, validation set, and testing set of the CSL-Daily-Event dataset contain 18401, 1077, and 1176 sign language videos, respectively. The CSL-Daily-Event dataset comprises 2,000 Chinese glosses and a Chinese vocabulary of 2,343 words.

We adopt BLEU~\cite{papineni2002bleu} and ROUGE-L~\cite{lin2004rouge} to assess the performance of sign language translation. 
BLEU mainly scores by comparing the matching degree of n-grams in the candidate translation with multiple reference translations. The higher the score, the greater the lexical similarity between the candidate translation and the reference translations. ROUGE-L calculates the recall rate based on the Longest Common Subsequence. It focuses more on evaluating the similarity in content coverage between the generated translation and the reference translations. For both of these metrics, a higher value indicates better performance.

\subsection{Implementation Details} 
To balance performance and efficiency, all images are resized to $224 \times 224$ during training and testing before being input to the network.
In the Mamba-based memory aggregation module, we stack two randomly initialized vision Mamba~\cite{Zhu2024Vision} layers to achieve efficient spatial feature aggregation. Meanwhile, in the graph-guided spatiotemporal fusion module, two HGCN~\cite{chami2019hyperbolic} layers are employed as the hypergraph learning network.
During training, we adopt the SGD~\cite{bottou2012stochastic} optimizer with an initial learning rate of 0.01, and employ a cosine annealing scheduler to adjust the learning rate over 200 epochs.
Our code is implemented using PyTorch~\cite{paszke2019pytorch} based on Python. All the experiments are conducted on a server with A800 GPUs. More details can be found in our source code.

\subsection{Comparison on Public SLT Datasets}

\noindent $\bullet$ \textbf{Results on Event-CSL Dataset.~}
As shown in Table~\ref{SL-Event_result}, we compare our method with other state-of-the-art SLT approaches on the Event-CSL dataset. Our model achieves superior performance, reaching ROUGE-L and BLEU-4 scores of 68.93 and 49.80, respectively.
% Among previous works, several gloss-based methods, such as Chen et al.~\cite{chen2022simple}, Sign-XmDA~\cite{ye2023cross}, GASLT~\cite{yin2023gloss}, and Joint-SLT~\cite{camgoz2020sign}, have demonstrated strong performance.
Compared with Sign-XmDA, which achieves the best results among gloss-based methods, our approach yields a substantial improvement of +37.41\% in BLEU-4. Furthermore, compared with recent gloss-free methods, like Sign2GPT~\cite{wong2024sign2gpt}, which leverages large language models for effective sign language translation, our model attains higher accuracy while maintaining a significantly smaller parameter size.
These results clearly demonstrate the effectiveness and efficiency of our approach on the Event-CSL dataset.

\noindent $\bullet$ \textbf{Results on EvSign Dataset.~}
% As shown in Table~\ref{EvSign_result},  a comparison is made between our method and the previous methods~\cite{camgoz2020sign, min2021visual, hu2023continuous, zhang2024evsign} for sign language translation. Among these gloss-based methods, Zhang et al.~\cite{zhang2024evsign} achieve excellent performance. To demonstrate the effectiveness of our approach, our method is compared with that of Zhang et al. Our method improves by +18.89 and +17.04 in ROUGE-L and BLEU-4, respectively, without any gloss annotations. These experimental results demonstrate the significant effectiveness of our method on the EvSign dataset. 
As shown in Table~\ref{EvSign_result}, we also evaluate our method on the EvSign dataset for a comprehensive comparison. It can be seen that our approach achieves the best overall performance among competing methods. Notably, by effectively integrating spatial detail representations, it outperforms GFSLT with a +2.32\% improvement in BLEU-4, further demonstrating the robustness and generalization of our method.

\noindent $\bullet$ \textbf{Results on PHOENIX-2014T-Event Dataset.} 
% These methods such as TSPNet~\cite{li2020tspnet}, Joint-SLT~\cite{camgoz2020sign}, Sign-XmDA~\cite{ye2023cross}, GASLT~\cite{yin2023gloss} and GFSLT~\cite{Zhou_2023_ICCV} exhibit excellent performance on the PHOENIX-2014T dataset. We make comparisons with them on the PHOENIX-2014T-Event dataset, as shown in Table~\ref{PHOENIX-2014T-Event_result}. Among the gloss-based methods, like Joint-SLT, Sign-XmDA, and GASLT, GASLT performs more prominently. Our method exceeds it by +12.61 and +7.71, respectively, in ROUGE-L and BLEU-4, even without the guidance of any sign language gloss. GFSLT achieves better results than the gloss-free methods. Compared to it, our method improves by +0.87 and +0.08 in the ROUGE-L and BLEU-4, respectively. 
As shown in Table~\ref{PHOENIX-2014T-Event_result}, we also report the results on the simulated PHOENIX-2014T-Event dataset. Among the gloss-based methods, such as Joint-SLT and Sign-XmDA, Sign-XmDA achieves the most competitive performance. In contrast, our method surpasses it by +10.77\% and +7.12\% in ROUGE-L and BLEU-4, respectively, without leveraging any sign language gloss annotations. Compared with GFSLT, our model achieves additional improvements of +1.53\% and +1.76\% on the ROUGE-L and BLEU-4 metrics, respectively. These experiments further demonstrate the generalization ability of our method on this dataset.

\noindent $\bullet$ \textbf{Results on CSL-Daily-Event Dataset.~}
% CSL-Daily-Event is an event simulation dataset that originates from the CSL-Daily sign language dataset. As shown in Table~\ref{CSL-Daily-Event_result}, we report multiple SOTA methods of SLT on the CSL-Daily-Event dataset. For the gloss-based methods, our method improves by +9.64 and +4.88, respectively, on ROUGE-L and BLEU-4 compared to GASLT. For the gloss-free methods, our method improves by +2.27 and +0.92, respectively, on ROUGE-L and BLEU-4 compared to GFSLT. The results of these experiments demonstrate that our method can also achieve good performance on the event simulation dataset.
As shown in Table~\ref{CSL-Daily-Event_result}, we report the results on the simulated CSL-Daily-Event dataset. For the gloss-based methods, our method improves by +13.22\% and +4.22\%, respectively, in ROUGE-L and BLEU-4 compared to Sign-XmDA. For the gloss-free methods, our method improves by +6.25\% and +0.72\%, respectively, compared to GFSLT. The results of these experiments demonstrate that our method can also achieve good performance on the event simulation dataset.

\subsection{Ablation Study} 
\noindent $\bullet$ \textbf{Component Analysis on Event-CSL.}  
% In order to demonstrate the validity of our proposed method, we separately assess the impact of each module on the experimental results. As shown in Table~\ref{SL-Event_result}, we conduct the component analysis on the Event-CSL dataset. We observe that when only using CNN as the visual backbone network, it shows excellent performance, indicating that it can capture the local spatial details of the video frames. Then, we introduce Mamba and construct a hybrid CNN-Mamba as the visual backbone network. Compared with CNN, the hybrid CNN-Mamba brings performance improvements in each metric, among which ROUGE-L, BLEU-1, BLEU-2, BLEU-3, and BLEU-4 increase by +0.30, +0.50, +0.41, +0.20, and +0.02, respectively. These experimental results prove that Mamba effectively models the long-range global temporal relations.
To demonstrate the effectiveness of our method, we evaluate the contribution of each module on the Event-CSL dataset, as shown in Table~\ref{CAResults}.
We first compare the temporal and spatial branches, and find that temporal features outperform spatial ones by a large margin, as sign language translation relies heavily on temporal continuity. 
% Although the spatial branch enhances spatial details, it is less effective in modeling temporal dependencies. 
Combining spatial details with temporal representations allows the model to better capture fine-grained motion and contextual cues.
We then examine the impact of the Mamba layer in the spatial branch. Replacing it with simple MLPs results in a 0.6\% drop in accuracy, confirming the benefit of the Mamba-based design. Finally, the fourth experiment validates the contribution of the proposed Graph-guided Spatiotemporal Fusion (GSTF) module. Overall, these results demonstrate the effectiveness of all components in our framework.

% \begin{table}
% \center
% \small   
% \caption{Component Analysis on the Event-CSL. } 
% \label{Event-CSL_Component}
% \begin{tabular}{l|cc|c|c|c|c|c}
% \hline \toprule [0.5 pt]
% No. &CNN &Mamba &R &B1 &B2 &B3 &B4 \\ 
% \hline
% 1  &\cmark  &  &67.72 &69.21 &60.94 &54.20 &48.68\\
% 2  &\cmark  &\cmark  &68.02 &69.71 &61.35 &54.40 &48.70 \\ 
% \hline \toprule [0.5 pt]
% \end{tabular}
% \end{table}

\begin{table}
\center
\small    
\caption{Analysis of \textbf{Input Frames}, \textbf{Input Resolutions}, and \textbf{Different Spatiotemporal Fusion Methods}. R represents ROUGE-L, and B-1, B-2, B-3, B-4 represent BLEU-1, BLEU-2, BLEU-3, and BLEU-4, respectively.}
\label{Ablation_Studies}
\resizebox{\columnwidth}{!}{ 
\begin{tabular}{c|ccccc}
\hline \toprule [0.5 pt]
\rowcolor{gray!20}
\textbf{\# Input Frames} &\textbf{R}  &\textbf{B-1} &\textbf{B-2} &\textbf{B-3} &\textbf{B-4} \\
\hline
L / 8  &29.93 &33.36 &22.01 &14.89 &10.62  \\ 
L / 4  &48.26 &50.20 &40.37 &33.23 &27.92  \\ 
L / 2  &59.69 &61.67 &52.52 &45.44 &39.80   \\ 
L  &\textbf{68.93}  &\textbf{70.40} &\textbf{62.09}	&\textbf{55.34} 	 	&\textbf{49.80}  \\ 
\hline 
\rowcolor{gray!20}
\textbf{\# Input Resolutions} &\textbf{R}  &\textbf{B-1} &\textbf{B-2} &\textbf{B-3} &\textbf{B-4} \\
\hline
$112 \times 112$  &48.04 &50.12 &42.16	&36.28 	&31.72 \\ 
$256 \times 256$  &68.52 &69.96 &61.54 &54.67 &49.06   \\ 
$320 \times 320$  &68.88 &70.38 &62.03 &55.24 &49.68   \\ 
$224 \times 224$  &\textbf{68.93}  &\textbf{70.40} &\textbf{62.09}	&\textbf{55.34} &\textbf{49.80} \\ 
\rowcolor{gray!20}
\textbf{Fusion Methods} &\textbf{R}  &\textbf{B-1} &\textbf{B-2} &\textbf{B-3} &\textbf{B-4} \\
\hline
Concatenate  &65.79 &67.32 &59.09 &52.43 &47.00  \\ 
Cross-Attention  &67.65 	&69.27 	&60.76 	&53.98 	  &48.84  \\ 
Ours \emph{w/o} Graph        &68.81 &69.82 &61.45 &54.65 &49.07   \\ 
Ours (GSTF)  &\textbf{68.93}  &\textbf{70.40} &\textbf{62.09}	&\textbf{55.34} &\textbf{49.80}   \\ 
\hline \toprule [0.5 pt]
\end{tabular}
}
\end{table}

\begin{table}
\center
\small    
\caption{Efficiency Analysis on Event-CSL dataset.}
\label{Efficiency_Analysis}
% \resizebox{\textwidth}{!}{ 
\begin{tabular}{c|c|c|c}
\hline \toprule [0.5 pt]
\textbf{Algorithm}  &\textbf{\# Params} & \textbf{\# FLOPs}  & \textbf{\# Speed} \\
\hline
SignCL~\cite{ye2024improving}   &115.5M  &266.5G &175 ms/video \\ 
Sign2GPT~\cite{wong2024sign2gpt}    &1771.1M  &490.5G &247 ms/video \\ 
EvSLT (Ours)  &138.2M  &252.0G  &86 ms/video \\ 
% 06 &Ours    &117.21M  &1654M  &1.81s  \\ 
% \hline
% \#06 &\textbf{GFSLT-VLP}~\cite{Zhou_2023_ICCV}    &ICCV2023  &CNN+Transformer  &35.53	&36.65 	&26.64 	&20.69 	&16.85   \\ 
\hline \toprule [0.5 pt] 
\end{tabular}
% }
\end{table}

\noindent $\bullet$ \textbf{Analysis of the Number of Input Event Frames.~} 
The number of input frames in sign language videos significantly affects both the model’s performance and its computational resource usage. To investigate this, we set four different frame sampling ratios: L/8, L/4, L/2, and L, where L denotes the total number of frames in each sign language video.
As shown in Table~\ref{Ablation_Studies}, we compare the model’s performance under these four frame settings on the Event-CSL dataset. The results demonstrate that as the number of frames increases, the model performance consistently improves. When all the video frames (L) are used, the performance is several times higher than that obtained using only L/8 frames. This indicates that increasing the number of input event frames enriches the temporal context, which in turn improves translation quality.

\noindent $\bullet$ \textbf{Analysis of Different Resolutions of Event Frames.~}  
% 1280*720， 640*320， 224*224 ？ 
As shown in Table~\ref{Ablation_Studies}, we conducted experiments using four different input resolutions for event frames: $112\times112$, $224\times224$, $256\times256$, and $320\times320$. It can be observed that the best performance is achieved at a resolution of $224\times224$. Larger resolutions, on the other hand, lead to performance degradation. We attribute this to the fact that high-resolution event inputs often introduce more redundant noise, and existing models are not yet well adapted to high-resolution event data. Nevertheless, we believe that high-definition event cameras still hold great potential in scenarios requiring detailed contour capture or larger receptive fields for tracking fast-moving targets. Existing high-resolution event datasets, such as EventVOT~\cite{wang2024eventvot}, EvDET200K~\cite{wang2025object}, and Celex-HAR~\cite{wang2024eventhar}, together with our proposed Event-CSL, will contribute to advancing research in this community.

\begin{figure*}
    \centering
    \includegraphics[width=7in]{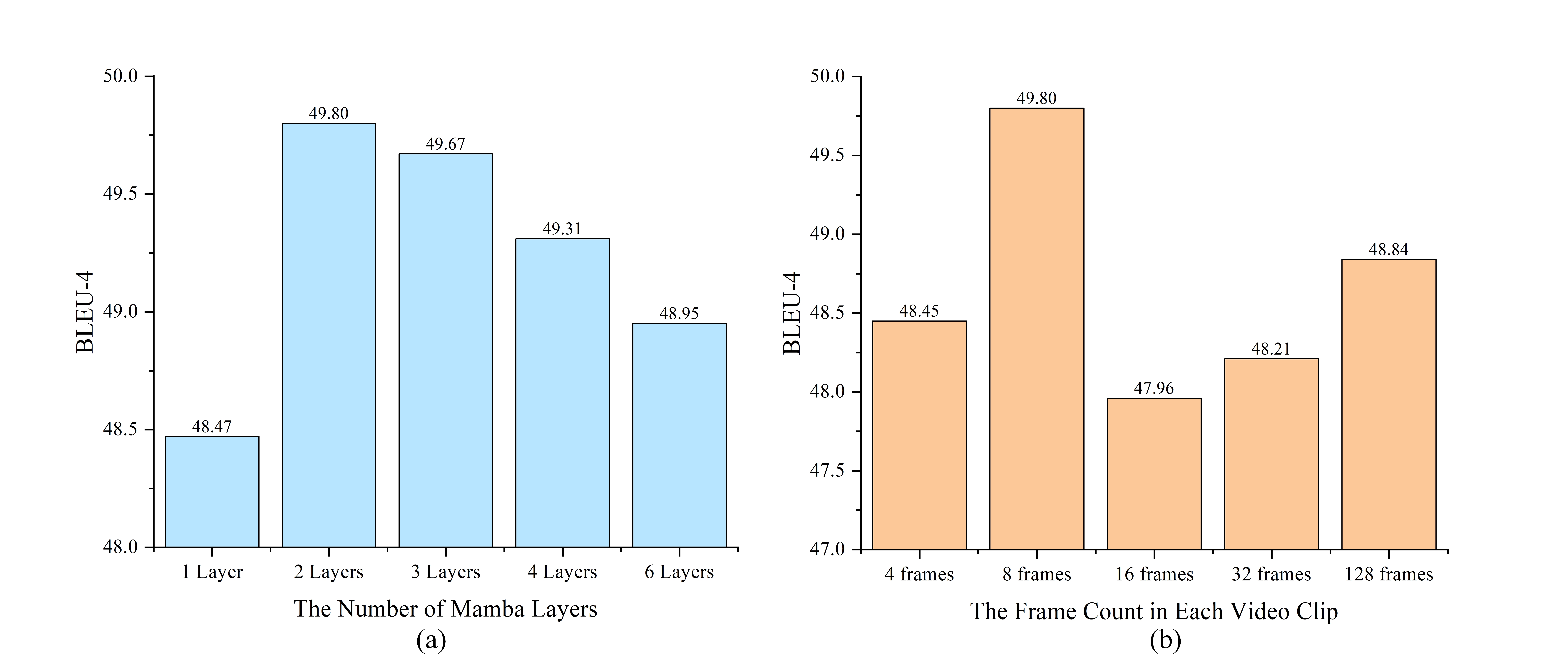}
    \caption{(a) Comparison with different numbers of Mamba layers. (b) Comparison with different frame counts in each video clip.}
    \label{fig:ablation_studies}
\end{figure*}

\begin{figure*}
    \centering
    \includegraphics[width=6.5in]{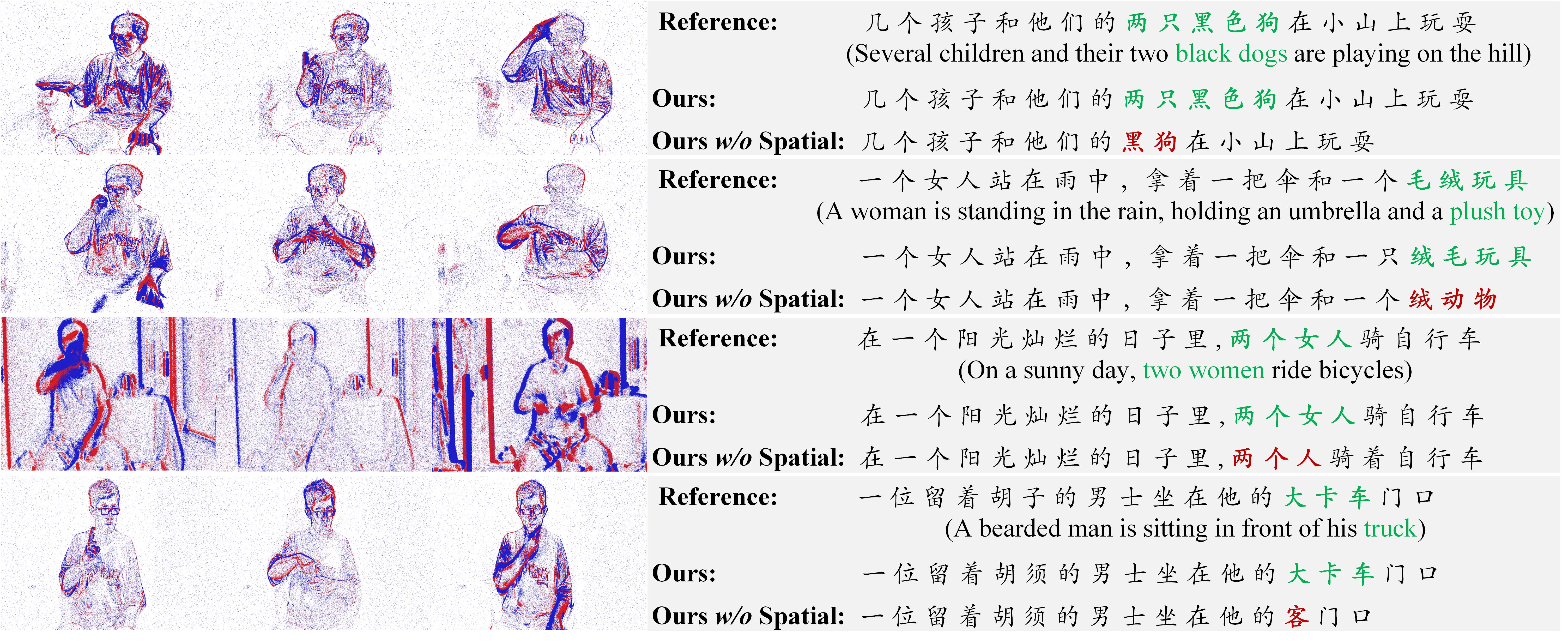}
    \caption{A qualitative comparison of the generated sign language translation texts.}
    \label{fig:quality_visual}
\end{figure*}

\noindent $\bullet$ \textbf{Analysis of Different Spatiotemporal Fusion Methods.~}
The graph-guided spatiotemporal fusion (GSTF) module is a key component of our framework. In Table~\ref{Ablation_Studies}, we compare it with other feature-level fusion methods. The results show that our spatiotemporal fusion approach outperforms alternative fusion strategies. Moreover, when the hypergraph modeling strategy is removed (Ours \emph{w/o} Graph), the performance drops significantly, demonstrating the effectiveness of using a hypergraph to capture relationships between video clips.

\noindent $\bullet$ \textbf{Analysis of the Depth of Mamba Network.~} 
As shown in Fig.~\ref{fig:ablation_studies} (a), we compare the effect of different numbers of Mamba layers on the final performance. The results show that two stacked Mamba layers yield the optimal configuration. When only a single Mamba layer is used, the model accuracy decreases by 1.33\%, indicating that the model has limited capacity to capture the complex spatiotemporal dependencies across video clips. In contrast, further increasing the number of Mamba layers also leads to performance degradation. Therefore, a two-layer Mamba provides a good trade-off between accuracy and efficiency, resulting in the best overall performance.

\noindent $\bullet$ \textbf{Analysis of the Frame Count in Each Video Clip.~} 
As shown in Fig.~\ref{fig:ablation_studies} (b), we analyze the number of frames contained in each video clip of the Event-CSL dataset. It can be observed that both too few and too many frames result in performance degradation. When the number of frames is too small, the model receives insufficient contextual information to capture the complete motion dynamics and contextual semantics of sign language. In contrast, an excessive number of frames introduces redundant or irrelevant visual information, increasing the model’s learning burden and even leading to overfitting. Therefore, we choose to include a fixed number of 8 frames per video clip to achieve the best performance.

\subsection{Efficiency Analysis} 
Table~\ref{CAResults} presents a component-wise performance analysis and the associated model parameters. The results show that our framework achieves notable performance gains with only a marginal increase in model parameters. By integrating spatial feature aggregation and spatiotemporal fusion, it achieves a balanced trade-off between complexity and accuracy.
Furthermore, as shown in Table~\ref{Efficiency_Analysis}, we compare our model with recent methods on parameters, FLOPs, and speed. Our model maintains low computational cost and delivers the fastest translation speed (approximately 86 ms per video). Notably, our method utilizes a unified single-stage framework, achieving higher translation accuracy while remaining computationally efficient.

\subsection{Visualization} 
% In this section, we report the qualitative results of the sign language translations generated by our model. 
% As shown in Fig.~\ref{fig:quality_visual}, we compare the generated sentences with the reference sentences and display the corresponding event sign language videos, where the red areas represent the differences between the generated sentences and the reference sentences. For the first row, we observe that the sentence generated by our model is exactly the same as the reference sentence. 
% For the second row, we find that the generated sentence has one more dynamic auxiliary word compared to the reference sentence. 
% For the third and fourth rows, it is not difficult to see that the model understands the content of the sign language videos and makes synonyms replacements for some conjunctions and verbs when generating the sentences, but the meaning remains the same. 
% For the last row, our model extracts the information of nouns but expands and modifies the nouns. 
% We believe that generating completely consistent nouns is also very challenging. 
% Overall, the quality of the sign language translations generated by our model is reliable. 
% Although most of the generated sentences are not exactly the same as the reference sentences, the wording is similar, and the overall meaning expressed is basically the same.
In this section, we present a qualitative analysis of the sign language translation results. As shown in the Fig.~\ref{fig:quality_visual}, \textbf{Reference} denotes the ground-truth translation sentence corresponding to the video, \textbf{Ours} represents the translation result generated by our model, and \textbf{Ours \emph{w/o} spatial} indicates the result obtained using only the temporal branch, without spatial detail features. The red text highlights incorrectly translated keywords, while the green text shows the correct translations of those keywords. It can be observed that, after removing spatial detail features, the model loses its ability to perceive and translate fine-grained sign details, e.g., “\begin{CJK}{UTF8}{gbsn}两只黑色狗\end{CJK} (two black dogs)” instead of simply “\begin{CJK}{UTF8}{gbsn}黑狗\end{CJK} (black dog)”. This demonstrates that our proposed spatiotemporal fusion method effectively enhances the model’s ability to translate detailed components.

% \subsection{Limitation Analysis}  
% Although our newly proposed SLT framework works well on multiple benchmark datasets, we believe this model can still be improved in the following aspects: 
% 1). We transform the event streams into event frames and resize them into $224 \times 224$. Designing novel architectures to better leverage high-definition event data remains an interesting direction for future research. 
% 2). We adopt a relatively lightweight language model (mBART) for sign language generation due to the limited computational resources. We believe a larger language model can bring us a much better performance. 
% We leave these two issues as our future work. 

\section{Limitation Analysis}  
% Our newly proposed sign language translation framework works well on multiple benchmark datasets, however, we believe this model can still be improved. We adopt a relatively lightweight large language model (LLM) for sign language generation due to the limited computational resources. We believe a larger language model can bring us a much better performance. We leave the issues as our future works. 
Although our newly proposed SLT framework works well on multiple benchmark datasets, we believe this model can still be improved in the following aspects: 
1). We transform the event streams into event frames and resize them into $224 \times 224$. Designing novel architectures to better leverage high-definition event data remains an interesting direction for future research. 
2). We adopt a relatively lightweight language model (mBART) for sign language generation due to the limited computational resources. We believe a larger language model can bring us a much better performance. 
We leave these two issues as our future work.

\section{Conclusion}  
% In this study, we have introduced a novel approach to Sign Language Translation (SLT) using high-definition event streams, addressing critical limitations of traditional SLT systems such as sensitivity to lighting conditions, rapid hand movements, and privacy concerns. The proposed Event-CSL dataset, comprising 14,827 videos and covering a diverse range of scenarios, significantly advances the field by providing a comprehensive resource for research and development in SLT. Our benchmarking of existing methods and the introduction of a new baseline that utilizes the Mamba model for enhanced temporal feature integration demonstrate the potential for improved translation outcomes in SLT. With the commitment to release both the Event-CSL dataset and the source code, this work paves the way for future research and the development of more effective and privacy-conscious SLT technologies.
In this paper, we propose Event-CSL, the first large-scale, high-definition sign language translation dataset captured using an event camera, aiming to bridge the data scarcity gap in this field. The dataset consists of 14,827 high-definition sign language videos with a resolution of $1280 \times 720$, along with 14,821 glosses and 2,544 Chinese words in the text vocabulary. We believe that Event-CSL will serve as a valuable resource for advancing research in event-based sign language translation.
Furthermore, we propose a unified spatiotemporal fusion framework for the sign language translation task. To enhance spatial detail representation, we first introduce a Mamba-based memory aggregation module to compress and aggregate video clip features. Then, a graph-guided spatiotemporal fusion module leverages the compressed clip features as a bridge to effectively fuse spatial features with temporal representations, enabling more precise capture of sign language details.
We hope that the proposed dataset and model will contribute to the advancement of this field and enable deaf individuals to communicate more freely.
% In future work, we plan to collect more high-definition sign language translation datasets and explore self-supervised pretraining strategies to develop a more powerful and generalizable sign language translation model.

{
    \small
    \bibliographystyle{ieeenat_fullname}
    \bibliography{reference}
}

\end{document}